\documentclass[final,1p,times]{elsarticle}

\usepackage[utf8]{inputenc}
\usepackage[T1]{fontenc}
\usepackage{makecell} 
\usepackage{booktabs} 
\usepackage{array}    
\usepackage{placeins}
\usepackage{subcaption}
\usepackage{algpseudocode}
\usepackage{algorithm}
\usepackage{fancyhdr}
\usepackage{comment}
\usepackage{xcolor}

\usepackage{tikz,graphicx}
\usepackage{multirow}
\usepackage{amsfonts,amsmath,amssymb}
\usepackage{mathtools}
\usepackage{mathrsfs}
\usepackage{upgreek}
\usepackage{lineno}

\begin{document}

\begin{frontmatter}

\title{Diffusion Model-Based Data Assimilation for Real-World Energy Consumption Forecasting}

\author[math]{Ruoyu Hu}
\author[cs]{Dahai Yu}
\author[math]{Feng Bao}
\author[cs]{Guang Wang}
\author[ornl]{Guannan Zhang}

\affiliation[math]{organization={Department of Mathematics, Florida State University},
            city={Tallahassee},
            postcode={32306},
            state={Florida},
            country={USA}}
\affiliation[cs]{organization={Department of Computer Science, Florida State University},
            city={Tallahassee},
            postcode={32306},
            state={Florida},
            country={USA}}
\affiliation[ornl]{organization={Computer Science and Mathematics Division, Oak Ridge National Laboratory},
            city={Oak Ridge},
            postcode={37831},
            state={Tennessee},
            country={USA}}
\begin{abstract}
Accurate estimation and forecasting of energy consumption are important for power-system operation, planning, and demand-side management. In practice, however, complete and timely measurements may not always be available, and the observed data can be partial, noisy, or delayed. This motivates the use of learned forecasting models for predicting the evolving consumption state, together with data assimilation methods for sequential forecast correction. In this work, we study a high-dimensional data assimilation problem for real energy-consumption data. The forward prediction is supplied by a pretrained black-box spatio-temporal forecasting model, which is treated as the state propagator in the filtering procedure. We employ the Ensemble Score Filter (EnSF) to assimilate partial and noisy observations and to correct the forecast trajectory over time. The EnSF uses score-based diffusion models to approximate filtering distributions and avoids retraining neural-network score models during assimilation by using a closed-form score representation and Monte Carlo approximation. Numerical experiments demonstrate that open-loop propagation of the learned forecasting model can become unreliable over long horizons, while EnSF-based correction substantially improves state estimation. Comparisons with the Ensemble Kalman Filter (EnKF) further show that EnSF provides stronger correction under the nonlinear observation setting considered in this work.
\end{abstract}


\end{frontmatter}

\section{Introduction}

Accurate energy-consumption forecasting is an important component of power-system operation, planning, and demand-side management. Electricity demand has continued to grow rapidly in recent years. The International Energy Agency reports that global electricity demand increased by $4.3\%$ in 2024 and is expected to grow at close to $4\%$ annually through 2027, driven by industrial activity, air conditioning, electrification, and data-center expansion~\cite{IEA_Electricity_2025}. Reliable load forecasting is therefore essential for generation scheduling, resource allocation, infrastructure planning, and operational reliability~\cite{NREL_Load_Forecasting_2023,Hong_Fan_2016}. However, real energy-consumption data often exhibit strong temporal variability, user-level heterogeneity, nonlinear dependence on external conditions, and noisy or incomplete measurements. Moreover, in practical forecasting settings, complete and timely measurements of the current consumption state may not always be available. These features make accurate state estimation and long-horizon prediction challenging, especially for high-dimensional user-level consumption trajectories~\cite{Akhtar_2023_STLF}.

Recent advances in machine learning have provided powerful tools for energy-consumption forecasting. In particular, deep spatio-temporal models can learn temporal dependence from historical consumption records while also extracting cross-user or cross-location dependence from high-dimensional data. Compared with classical statistical forecasting models, these learned models are better suited for capturing nonlinear temporal patterns and heterogeneous user-level behavior. In this work, the forward model is a pretrained spatio-temporal forecasting model developed for real-data energy-consumption prediction. The model takes historical consumption information as input and predicts the next-step energy-consumption state for all users in the dataset.

The pretrained forecasting model used in this paper is treated as a learned black-box state propagator. Its parameters are fixed throughout the data assimilation experiments. The purpose of the present paper is not to retrain or modify the forecasting model, but to investigate how data assimilation can be used to correct the forecast trajectory generated by such a pretrained model. The model architecture, training procedure, and dataset construction are described in the following sections.

Although modern machine-learning forecasting models can capture complex spatio-temporal patterns from historical data, their predictions may still deviate from the true consumption trajectory when the model is recursively propagated over long horizons. This issue is not only a forecasting-model problem, but also a state-estimation problem: the latent energy-consumption state evolves dynamically, while the available observations may be partial, noisy, nonlinear, or delayed. Therefore, after a learned forward model has been constructed, a natural next step is to combine the model forecast with observational information collected over time. Data assimilation provides a systematic framework for this model-data fusion task~\cite{law2015data}.

Data assimilation plays a crucial role in bridging the gap between model predictions and observational data. In sequential forecasting problems, a forward model provides a prediction of the system state, while observations provide incomplete and noisy information about the true state. By systematically integrating these two sources of information, data assimilation methods refine state estimates and account for uncertainties arising from both the model and the data. This is particularly important for high-dimensional real-data forecasting problems, where direct reliance on the forward model alone may lead to inaccurate long-horizon prediction~\cite{carrassi2018data}.

The primary mathematical framework for addressing data assimilation problems is \textit{optimal filtering}. The goal of the optimal filtering problem is to approximate the conditional probability density function of the state process given the observational data, which is referred to as the \textit{filtering distribution}~\cite{Bao_Zakaid_2015,vanLeeuwen_2009}. The corresponding conditional expectation provides the optimal state estimate. Bayesian inference plays a central role in solving optimal filtering problems, with Kalman-type filters being a prominent example. For linear state dynamics and linear observations under Gaussian assumptions, the Kalman filter provides an analytical solution.

For nonlinear systems, ensemble Kalman filters (EnKF) were developed as computationally efficient ensemble-based approximations~\cite{evensen1994sequential,houtekamer1998data}. EnKF has been widely used in high-dimensional applications because it avoids explicit propagation of the full covariance matrix and instead represents uncertainty through an ensemble of forecast samples. However, the EnKF update is based on Gaussian-type approximations of the filtering distribution, which may become restrictive when the dynamics, observations, or posterior distributions are strongly nonlinear or non-Gaussian. In such cases, the filtering distribution may contain structures that cannot be accurately represented by only its ensemble mean and covariance.

Beyond EnKF, several methods have been proposed for nonlinear optimal filtering, including particle filters~\cite{andrieu2010particle,chorin2009implicit,gordon1993novel,kang2018improved,pitt1999filtering,snyder2008obstacles,bao2014hybrid}, Zakai filters~\cite{zakai1969optimal,liang2024online}, and methods based on stochastic partial differential equations~\cite{bao2021data,bao2015forward,bao2016first,bao2017adaptive,hu2002approximation,gobet2006discretization}. Particle filters, also known as sequential Monte Carlo methods, are well suited for representing non-Gaussian filtering distributions in moderate-dimensional problems. However, their performance can deteriorate in high-dimensional settings due to weight degeneracy and the difficulty of high-dimensional sampling in Bayesian inference. SPDE-based filtering methods provide a rigorous probabilistic formulation for nonlinear filtering, but their numerical implementation can become computationally expensive as the state dimension increases~\cite{Bao_Zakaid_2015,zhang2008grid,cai2020learning,dhariwal2021diffusion}. These challenges motivate filtering methods that can handle nonlinear and non-Gaussian state estimation while remaining computationally feasible in high-dimensional systems.

In this work, we employ the Ensemble Score Filter (EnSF) as a score-based data assimilation method for correcting forecasts generated by a pretrained energy-consumption forecasting model. The EnSF is built upon score-based diffusion models~\cite{NEURIPS2021_49ad23d1,NEURIPS2020_4c5bcfec,NEURIPS2019_3001ef25,song2021scorebased}, which characterize probability distributions through their score functions and generate samples through reverse-time diffusion processes. Traditional diffusion-based filtering methods~\cite{bao2023scorebased} estimate score functions with neural networks, which may require repeated training and large storage costs in sequential high-dimensional filtering problems. In contrast, EnSF avoids training a score network. It directly approximates the score function using a closed-form expression and mini-batch Monte Carlo estimators, and incorporates observational information through an analytical update mechanism. This provides a practical score-based filtering framework for high-dimensional nonlinear data assimilation~\cite{EnSF_2023,Bao_SC_2024,Inpainting_2024,Bao_MWR_2024}.

The objective of this paper is to investigate EnSF-based forecast correction for real-data energy-consumption prediction. The pretrained forecasting model is treated as a black-box forward propagator, while the filtering procedure sequentially assimilates partial and noisy observations to correct the forecast trajectory. The main point of the proposed framework is that the learned forecasting model supplies dynamical prediction when complete current-state information is unavailable, while data assimilation prevents the forecast from relying solely on open-loop model propagation. We compare the open-loop forecast, EnSF-corrected forecast, and EnKF-corrected forecast under different observation settings. The numerical results show that data assimilation substantially improves long-horizon state estimation and that EnSF provides stronger correction than EnKF in the nonlinear observation setting considered in this work.

The rest of this paper is organized as follows. Section~\ref{sec:model} introduces the pretrained energy-consumption forecasting model used as the forward propagator. Section~\ref{sec:state_estimation_da} formulates the data assimilation problem and presents the Ensemble Score Filter methodology. Section~\ref{sec:numerical_experiments} presents numerical experiments on real energy-consumption data. These experiments evaluate the behavior of the pretrained forecasting models, demonstrate the need for data assimilation in long-horizon forecast correction, and compare EnSF with EnKF under partial and nonlinear observation settings. Finally, the conclusion summarizes the main findings and discusses possible directions for future work.

\section{Pretrained energy-consumption forecasting model}
\label{sec:model}

The forward model used in this work is a pretrained spatio-temporal large language model (STLLM) designed for energy-consumption forecasting. The model adapts several components commonly used in modern large language models, including RMSNorm, rotary positional embeddings (RoPE), and SwiGLU feed-forward networks, to the spatio-temporal forecasting setting. Its main architectural feature is a factored attention structure that separates temporal dependency modeling from spatial dependency modeling.

\subsection{Input-output formulation}
\label{sec:model_input_output}

The input to STLLM is a four-dimensional tensor
\begin{equation}
    \mathbf{X}^{\mathrm{in}}
    \in
    \mathbb{R}^{B\times T\times N\times D_{in}},
    \label{eq:stllm_input_tensor}
\end{equation}
where $B$ is the batch size, $T$ is the historical sequence length, $N$ is the number of users or spatial nodes, and $D_{in}$ is the input feature dimension. The model outputs a prediction tensor
\begin{equation}
    \widehat{\mathbf{Z}}
    \in
    \mathbb{R}^{B\times H\times N\times D_{out}},
    \label{eq:stllm_output_tensor}
\end{equation}
where $H$ is the forecast horizon and $D_{out}$ is the output feature dimension. The notation $\widehat{\mathbf{Z}}$ is used here for the model prediction tensor in order to distinguish it from the observation variable $Y$ used in the data assimilation formulation.

Abstractly, let
\begin{equation}
    W_n
    =
    \left(
        X_{t_{n-T+1}},
        X_{t_{n-T+2}},
        \ldots,
        X_{t_n}
    \right)
    \label{eq:stllm_input_window}
\end{equation}
denote the historical input window at time $t_n$. The pretrained model defines the learned one-step forecasting map
\begin{equation}
    \widehat{X}_{t_{n+1}}
    =
    f_{\theta}(W_n),
    \label{eq:stllm_forward_map_window}
\end{equation}
where $\theta$ denotes the fixed pretrained model parameters. For notational simplicity in the data-assimilation formulation, we write the learned forecast map as $f_{\theta}$. Throughout this paper, the parameters $\theta$ are fixed. The forecasting model is treated as a black-box state propagator, and the role of data assimilation is to correct the forecasted state using observational information rather than to retrain the model.

\subsection{Architecture summary}
\label{sec:stllm_architecture}

The architecture consists of an input embedding layer, stacked STLLM blocks, and an output projection layer. The embedding layer maps the raw input feature into a latent space of dimension $d$ and adds a learnable node embedding to encode the identity of each user or spatial node. This allows the model to distinguish different users without requiring an explicit graph adjacency matrix as a mandatory input.

Each STLLM block contains three main sub-layers with residual connections: temporal attention, spatial attention, and a SwiGLU feed-forward network. Temporal attention is applied along the historical time dimension for each node, and RoPE is used to encode relative temporal position information. Spatial attention is applied across users at each time step, allowing the model to learn inter-user dependency patterns adaptively. The feed-forward component uses a SwiGLU activation to improve the expressive capability of the hidden representation. All sub-layers use RMSNorm and pre-normalization residual connections.

After the final STLLM block, the hidden representation at the last input time step is selected and projected into the prediction space. This produces the next-step forecast used in the subsequent data-assimilation procedure.

\begin{table}[ht]
\caption{Summary of the STLLM architectural components.}
\label{tab:stllm_architecture_summary}
\centering
\small
\begin{tabular}{lcc}
\toprule
\textbf{Component} & \textbf{Standard Transformer} & \textbf{STLLM} \\
\midrule
Normalization & LayerNorm & RMSNorm \\
Position encoding & Sinusoidal / learned & RoPE \\
Feed-forward activation & ReLU / GELU & SwiGLU \\
Attention structure & Joint attention & Factored temporal-spatial attention \\
Normalization placement & Post-norm & Pre-norm \\
Spatial modeling & Usually graph-dependent or implicit & Adaptive spatial attention \\
\bottomrule
\end{tabular}
\end{table}


\section{Estimation of states through data assimilation}
\label{sec:state_estimation_da}

\subsection{The data assimilation problem}
\label{sec:da_problem}

We formulate the forecast-correction task as a sequential data assimilation problem. Let $X_{t_n}\in\mathbb{R}^{d_x}$ denote the energy-consumption state at time $t_n$. The state dimension $d_x$ depends on the set of energy-consuming users included in the dataset. The pretrained forecasting model introduced above is used as the forward propagator. Since the model predicts the next state from a historical input window, the state evolution can be formulated as
\begin{equation}
    X_{t_{n+1}}
    =
    f_{\theta}(W_n,\eta_{t_n}),
    \qquad n=0,1,2,\ldots,
    \label{eq:state_model_energy}
\end{equation}
where $f_{\theta}$ denotes the learned black-box forecast model, $W_n$ is the historical input window defined in \eqref{eq:stllm_input_window}, and $\eta_{t_n}$ represents model error or stochastic uncertainty. For the $1$-to-$1$ model, the input window reduces to the single current state, while for the $4$-to-$1$ and $12$-to-$1$ models, $W_n$ contains the corresponding four-step or twelve-step state history.

The goal of the data assimilation procedure is to fuse observational information about $X$ with the forecast obtained from the forward model \eqref{eq:state_model_energy} in order to obtain an improved estimate of the state. We denote the observation at time $t_{n+1}$ by $Y_{t_{n+1}}$, given by
\begin{equation}
    Y_{t_{n+1}}
    =
    \mathcal{H}_{n+1}(X_{t_{n+1}})
    +
    \epsilon_{t_{n+1}},
    \label{eq:obs_model_energy}
\end{equation}
where $Y_{t_{n+1}}$ is the observation vector, $\mathcal{H}_{n+1}$ is the observation operator that may provide incomplete and indirect measurements of the state, and $\epsilon_{t_{n+1}}$ denotes observational noise.

With \eqref{eq:state_model_energy} and \eqref{eq:obs_model_energy}, the data assimilation problem becomes finding the best estimate of the latent state from all available observations. The optimal filtering estimate at time $t_{n+1}$ is given by
\begin{equation}
    \hat{X}_{t_{n+1}}
    :=
    \mathbb{E}\!\left[
    X_{t_{n+1}}
    \mid
    Y_{t_1:t_{n+1}}
    \right],
    \label{eq:optimal_estimate_energy}
\end{equation}
where $Y_{t_1:t_{n+1}}$ denotes the observational information collected from $t_1$ to $t_{n+1}$. Equivalently, the estimate is the conditional expectation of $X_{t_{n+1}}$ with respect to the $\sigma$-algebra generated by these observations. For nonlinear high-dimensional systems, this conditional distribution is generally unavailable in closed form. Instead, one works on approximating the conditional probability density function of the state,
$
    p(X_{t_{n+1}} \mid Y_{t_1:t_{n+1}}),
$
which is referred to as the filtering distribution.

The standard approach to solving the data assimilation problem is the Bayesian filtering framework. The recursive Bayesian filter consists of a prediction step and an update step. Suppose that the filtering distribution $p(X_{t_n}\mid Y_{t_1:t_n})$ has been approximated at time $t_n$. By using the Chapman--Kolmogorov formula, one propagates the state equation in \eqref{eq:state_model_energy} from $t_n$ to $t_{n+1}$ and obtains the prior filtering distribution
\begin{equation}
    p(X_{t_{n+1}}\mid Y_{t_1:t_n})
    =
    \int
    p(X_{t_{n+1}}\mid X_{t_n})
    p(X_{t_n}\mid Y_{t_1:t_n})
    \,dX_{t_n}.
    \label{eq:prediction_energy}
\end{equation}
Here, $p(X_{t_{n+1}}\mid X_{t_n})$ is the transition probability derived from the state dynamics introduced in \eqref{eq:state_model_energy}. The resulting prior filtering distribution $p(X_{t_{n+1}}\mid Y_{t_1:t_n})$ combines the observational information collected up to time $t_n$ with the model information propagated to time $t_{n+1}$.

After the new observation $Y_{t_{n+1}}$ becomes available, the update step incorporates it through Bayesian formula:
\begin{equation}
    p(X_{t_{n+1}}\mid Y_{t_1:t_{n+1}})
    \propto
    p(X_{t_{n+1}}\mid Y_{t_1:t_n})
    p(Y_{t_{n+1}}\mid X_{t_{n+1}}).
    \label{eq:update_energy}
\end{equation}
If the observational noise is Gaussian with covariance matrix $R_{n+1}$, the likelihood is
\begin{equation}
    p(Y_{t_{n+1}}\mid X_{t_{n+1}})
    \propto
    \exp\!\left[
    -\frac{1}{2}
    \left(
        \mathcal{H}_{n+1}(X_{t_{n+1}})-Y_{t_{n+1}}
    \right)^{\top}
    R_{n+1}^{-1}
    \left(
        \mathcal{H}_{n+1}(X_{t_{n+1}})-Y_{t_{n+1}}
    \right)
    \right].
    \label{eq:likelihood_energy}
\end{equation}

This Bayesian formulation provides the foundation for the filtering methods considered in this work. In the following section, we introduce the Ensemble Score Filter for solving the data assimilation problem by utilizing the concept of score-based diffusion models within the generative artificial intelligence framework.

\subsection{Score-based diffusion model and training-free score approximation}
\label{sec:score_diffusion_training_free}

We first introduce the score-based diffusion formulation that will be used to represent and sample from filtering distributions. The purpose of the diffusion construction is to transform a target distribution into a standard Gaussian distribution through a forward stochastic process, and then recover samples from the target distribution by solving the corresponding reverse-time process. A key quantity in this procedure is the score function, which stores the distributional information needed to guide the reverse-time sampling.

Let $Z_{\tau}\in \mathbb{R}^{d_x}$ be a diffusion process defined on the pseudo-time interval $\tau\in[0,1]$. The variable $\tau$ is independent of the physical time variable in the data assimilation problem. We consider the forward SDE
\begin{equation}
    dZ_{\tau}=b(\tau)Z_{\tau}\,d\tau+\sigma(\tau)\,dW_{\tau},
    \label{eq:score_forward_sde}
\end{equation}
where $W_{\tau}$ is a standard Brownian motion, and $b(\tau)$ and $\sigma(\tau)$ are the drift and diffusion coefficients, respectively. Following the score-based diffusion construction in \cite{song2021scorebased}, we choose
\begin{equation}
    b(\tau)=\frac{d\log \alpha_{\tau}}{d\tau},
    \qquad
    \sigma^2(\tau)=\frac{d\beta_{\tau}^2}{d\tau}
    -2\frac{d\log \alpha_{\tau}}{d\tau}\beta_{\tau}^2,
    \label{eq:score_coefficients}
\end{equation}
with $\alpha_{\tau}=1-\tau$ and $\beta_{\tau}^2=\tau$ for $\tau\in[0,1]$. Since \eqref{eq:score_forward_sde} is linear, this choice gives the conditional density
\begin{equation}
    Q_{\tau}(Z_{\tau}\mid Z_0)
    =
    \mathcal{N}\left(\alpha_{\tau}Z_0,\beta_{\tau}^2 I_{d_x}\right)
    \label{eq:score_conditional_density}
\end{equation}
for any fixed initial value $Z_0$. Hence, the forward diffusion maps the initial target distribution toward the standard Gaussian distribution as $\tau\to 1$.

To recover samples from the target distribution, one considers the reverse-time SDE
\begin{equation}
    dZ_{\tau}
    =
    \left[
        b(\tau)Z_{\tau}
        -
        \sigma^2(\tau)\mathcal{S}(Z_{\tau},\tau)
    \right]d\tau
    +
    \sigma(\tau)d\overleftarrow{W}_{\tau},
    \label{eq:score_reverse_sde}
\end{equation}
where $d\overleftarrow{W}_{\tau}$ denotes a backward It\^o stochastic integral \cite{SDE,BDSDE}. The score function is defined by
\begin{equation}
    \mathcal{S}(z,\tau):=\nabla_z\log Q_{\tau}(z),
    \label{eq:score_function}
\end{equation}
where $Q_{\tau}$ denotes the probability density of $Z_{\tau}$. If the score function is available, then the reverse-time process \eqref{eq:score_reverse_sde} transports samples from the standard Gaussian distribution at $\tau=1$ back to the target distribution at $\tau=0$ \cite{SF_2023}.

The main computational issue is the approximation of $\mathcal{S}(z,\tau)$. Let $Q_0$ be the target distribution at $\tau=0$. From \eqref{eq:score_conditional_density}, the marginal density of $Z_{\tau}$ is $Q_{\tau}(z)=\int_{\mathbb{R}^{d_x}}Q_{\tau}(z\mid z_0)Q_0(z_0)\,dz_0$. Therefore, the score function can be rewritten as
\begin{equation}
\begin{aligned}
    \mathcal{S}(z,\tau)
    &=
    \nabla_z\log
    \left(
        \int_{\mathbb{R}^{d_x}}
        Q_{\tau}(z\mid z_0)Q_0(z_0)\,dz_0
    \right)  \\
    &=
    \frac{
        \int_{\mathbb{R}^{d_x}}
        \nabla_z Q_{\tau}(z\mid z_0)Q_0(z_0)\,dz_0
    }{
        \int_{\mathbb{R}^{d_x}}
        Q_{\tau}(z\mid z'_0)Q_0(z'_0)\,dz'_0
    } \\
    &=
    \int_{\mathbb{R}^{d_x}}
    -
    \frac{z-\alpha_{\tau}z_0}{\beta_{\tau}^2}
    w_{\tau}(z,z_0)
    Q_0(z_0)\,dz_0,
\end{aligned}
\label{eq:score_integral_representation}
\end{equation}
where the weight function is
\begin{equation}
    w_{\tau}(z,z_0)
    :=
    \frac{
        Q_{\tau}(z\mid z_0)
    }{
        \int_{\mathbb{R}^{d_x}}
        Q_{\tau}(z\mid z'_0)Q_0(z'_0)\,dz'_0
    }.
    \label{eq:score_weight}
\end{equation}
The weight function satisfies $\int_{\mathbb{R}^{d_x}}w_{\tau}(z,z_0)Q_0(z_0)\,dz_0=1$. Equation~\eqref{eq:score_integral_representation} expresses the score function as an expectation with respect to the target distribution $Q_0$.

In standard score-based diffusion models, the score function is commonly approximated by training a neural network. For filtering problems, however, the target distribution changes recursively as new observations are assimilated, and retraining a score network at each assimilation step is computationally inefficient. Instead, we use a training-free Monte Carlo approximation of the score. Suppose that $\{z^{(m)}\}_{m=1}^{M}$ is an ensemble of samples from $Q_0$. Given a mini-batch $\{z^{(j)}\}_{j=1}^{J}$ with $J\leq M$, we approximate the score by
\begin{equation}
    \mathcal{S}(z,\tau)
    \approx
    \bar{\mathcal{S}}(z,\tau)
    :=
    \sum_{j=1}^{J}
    -
    \frac{z-\alpha_{\tau}z^{(j)}}{\beta_{\tau}^2}
    \bar{w}_{\tau}(z,z^{(j)}),
    \label{eq:score_mc_approx}
\end{equation}
where the empirical weight is
\begin{equation}
    \bar{w}_{\tau}(z,z^{(j)})
    :=
    \frac{
        Q_{\tau}(z\mid z^{(j)})
    }{
        \sum_{\ell=1}^{J}Q_{\tau}(z\mid z^{(\ell)})
    }.
    \label{eq:score_mc_weight}
\end{equation}
Thus, the weights are computed by normalizing the Gaussian transition density values $\{Q_{\tau}(z\mid z^{(j)})\}_{j=1}^{J}$. This approximation avoids neural-network training and evaluates the score directly from the available ensemble samples. In practice, the mini-batch can be a small subset of the full ensemble while still providing sufficient accuracy for filtering problems \cite{EnSF_2023}.

\subsection{The ensemble score filter for state estimation}
\label{sec:ensf_state_estimation}

We now introduce how the Ensemble Score Filter (EnSF) is used to solve the data assimilation problem formulated in Section~\ref{sec:da_problem}. The methodology of EnSF is to define score functions corresponding to the filtering distributions and approximate these scores using the training-free Monte Carlo construction introduced above. In the present work, the pretrained forecasting model $f_{\theta}$ is treated as the black-box forward propagator, and EnSF is used to update the forecasted state by assimilating observational data.

Assume that $\mathcal{S}_{n|n}$ is the approximated score function corresponding to the posterior filtering distribution $p(X_{t_n}\mid Y_{t_1:t_n})$. For a uniform discretization of the pseudo-time interval $[0,1]$, let
$$
0=\tau_0<\tau_1<\cdots<\tau_l<\cdots<\tau_L=1,
\qquad
\Delta\tau=\tau_{l+1}-\tau_l.
$$
Starting from Gaussian samples at $\tau=1$, we generate an ensemble of state samples $\{z_{n|n}^{(m)}\}_{m=1}^{M}$ for $p(X_{t_n}\mid Y_{t_1:t_n})$ by solving the reverse-time SDE with the Euler--Maruyama scheme
\begin{equation}
\begin{aligned}
    \bar{Z}_{\tau_l}^{(m)}
    =
    \bar{Z}_{\tau_{l+1}}^{(m)}
    -
    \left[
        b(\tau_{l+1})\bar{Z}_{\tau_{l+1}}^{(m)}
        -
        \sigma^2(\tau_{l+1})
        \mathcal{S}_{n|n}
        \left(
            \bar{Z}_{\tau_{l+1}}^{(m)},\tau_{l+1}
        \right)
    \right]\Delta\tau
    +
    \sigma(\tau_{l+1})\Delta\overleftarrow{W}_{\tau_{l+1}}^{(m)},
\end{aligned}
\label{eq:ensf_reverse_sde_discretized}
\end{equation}
for $l=L-1,\ldots,0$, with terminal condition $\bar{Z}_{\tau_L}^{(m)}\sim\mathcal{N}(0,I_{d_x})$. The generated samples at $\tau=0$ are identified with the posterior ensemble at time $t_n$, namely $z_{n|n}^{(m)}:=\bar{Z}_{\tau_0}^{(m)}$ for $m=1,\ldots,M$.

In the prediction step, these posterior ensemble samples are propagated through the pretrained forecasting model. Since the forecasting model uses a historical input window, we define the ensemble input window for the $m$-th ensemble member by
\begin{equation}
    W_n^{(m)}
    =
    \left(
        z_{n-T+1|n-T+1}^{(m)},
        z_{n-T+2|n-T+2}^{(m)},
        \ldots,
        z_{n|n}^{(m)}
    \right),
    \qquad m=1,\ldots,M,
    \label{eq:ensf_energy_ensemble_window}
\end{equation}
where $W_n^{(m)}$ represents the sequence of posterior ensemble states used as the input window for the $m$-th forecast. The predicted ensemble member is then computed by
\begin{equation}
    z_{n+1|n}^{(m)}
    =
    f_{\theta}
    \left(
        W_n^{(m)},\eta_{t_n}^{(m)}
    \right),
    \qquad
    m=1,\ldots,M.
    \label{eq:ensf_energy_prediction_sample}
\end{equation}
The predicted ensemble $\{z_{n+1|n}^{(m)}\}_{m=1}^{M}$ provides a sample approximation of the prior filtering distribution $p(X_{t_{n+1}}|Y_{t_1:t_n})$. After the posterior ensemble $\{z_{n+1|n+1}^{(m)}\}_{m=1}^{M}$ is obtained, the corresponding input window is updated by removing the oldest state and appending the newly corrected state.
The corresponding prior score is approximated by the Monte Carlo scheme
\begin{equation}
    \bar{\mathcal{S}}_{n+1|n}(z,\tau)
    :=
    \sum_{j=1}^{J}
    -
    \frac{
        z-\alpha_{\tau}z_{n+1|n}^{(j)}
    }{
        \beta_{\tau}^{2}
    }
    \bar{w}_{\tau}
    \left(
        z,
        z_{n+1|n}^{(j)}
    \right),
    \qquad \tau\in[0,1],
    \label{eq:ensf_energy_prior_score}
\end{equation}
where $\{z_{n+1|n}^{(j)}\}_{j=1}^{J}$ is a mini-batch selected from the predicted ensemble, with $J\leq M$. The empirical weight is given by
\begin{equation}
    \bar{w}_{\tau}
    \left(
        z,
        z_{n+1|n}^{(j)}
    \right)
    :=
    \frac{
        Q_{\tau}
        \left(
            z\mid z_{n+1|n}^{(j)}
        \right)
    }{
        \sum_{\ell=1}^{J}
        Q_{\tau}
        \left(
            z\mid z_{n+1|n}^{(\ell)}
        \right)
    },
    \label{eq:ensf_energy_prior_weight}
\end{equation}
where $Q_{\tau}$ is the Gaussian transition density of the forward diffusion process. We call $\bar{\mathcal{S}}_{n+1|n}$ the prior score associated with the prior filtering distribution.

To generate state samples that follow the posterior filtering distribution, we update the prior score by incorporating the new observation $Y_{t_{n+1}}$. According to Bayesian rule,
$$
p(X_{t_{n+1}}\mid Y_{t_1:t_{n+1}})
\propto
p(X_{t_{n+1}}\mid Y_{t_1:t_n})
p(Y_{t_{n+1}}\mid X_{t_{n+1}}).
$$
Taking the logarithmic gradient motivates the posterior score approximation
\begin{equation}
    \bar{\mathcal{S}}_{n+1|n+1}(z,\tau)
    =
    \bar{\mathcal{S}}_{n+1|n}(z,\tau)
    +
    g(\tau)
    \nabla_z
    \log p(Y_{t_{n+1}}\mid X_{t_{n+1}})(z),
    \label{eq:ensf_energy_posterior_score}
\end{equation}
where $p(Y_{t_{n+1}}\mid X_{t_{n+1}})$ is the likelihood function defined in \eqref{eq:likelihood_energy}. The function $g(\tau)$ is a damping function that controls how the observational information is incorporated in the diffusion domain. In the current EnSF framework, $g(\tau)$ is monotonically decreasing on $[0,1]$ and satisfies $g(0)=1$ and $g(1)=0$. This condition indicates that the likelihood information is fully incorporated at $\tau=0$, where the target posterior distribution is recovered, while it vanishes at $\tau=1$, where the reverse diffusion process is connected to the Gaussian reference distribution. In this work, we use
\begin{equation}
    g(\tau)=1-\tau.
    \label{eq:ensf_energy_damping_choice}
\end{equation}

For the Gaussian observation model in \eqref{eq:likelihood_energy}, the likelihood-gradient term in \eqref{eq:ensf_energy_posterior_score} can be written as
\begin{equation}
    \nabla_z
    \log p(Y_{t_{n+1}}\mid X_{t_{n+1}})(z)
    =
    -
    D\mathcal{H}_{n+1}(z)^{\top}
    R_{n+1}^{-1}
    \left(
        \mathcal{H}_{n+1}(z)-Y_{t_{n+1}}
    \right),
    \label{eq:ensf_energy_likelihood_gradient}
\end{equation}
where $D\mathcal{H}_{n+1}(z)$ is the Jacobian of the observation operator. For direct partial observations, this expression reduces to a masked linear correction. For nonlinear observations, such as the arctangent observation operator considered in our experiments, the Jacobian accounts for the nonlinear measurement map.

Finally, the posterior score $\bar{\mathcal{S}}_{n+1|n+1}$ is used in the reverse-time SDE to generate posterior ensemble samples $\{z_{n+1|n+1}^{(m)}\}_{m=1}^{M}$ that approximate $p(X_{t_{n+1}}\mid Y_{t_1:t_{n+1}})$. The filtered state estimate is then computed as the ensemble mean
\begin{equation}
    \widehat{X}_{t_{n+1}}
    =
    \frac{1}{M}
    \sum_{m=1}^{M}
    z_{n+1|n+1}^{(m)}.
    \label{eq:ensf_energy_state_estimate}
\end{equation}

Thus, one EnSF assimilation step proceeds as follows. First, posterior samples at time $t_n$ are generated from the current posterior score model. Second, these samples are propagated through the pretrained forecasting model to form the prior ensemble. Third, the prior ensemble is used to construct the Monte Carlo approximation of the prior score. Fourth, the new observation is incorporated through the likelihood-gradient correction to obtain the posterior score. Finally, the reverse-time diffusion process driven by the posterior score generates the updated ensemble at time $t_{n+1}$.

\section{Numerical experiments}
\label{sec:numerical_experiments}

\subsection{Dataset, preprocessing, and model configurations}
\label{sec:data_model_setup}

The numerical experiments are conducted on a real energy-consumption dataset containing hourly records for $N=5{,}000$ utility users in Tallahassee, Florida. Each full yearly record contains $8{,}736$ hourly time steps, corresponding to $364$ days with $24$ hourly records per day. Each user has one recorded feature, namely energy consumption measured in kWh. An adjacency matrix of size $5{,}000\times 5{,}000$ describing spatial proximity between users is also available in the dataset, although the STLLM architecture can learn spatial dependence through attention without relying explicitly on a fixed graph structure. The numerical examples reported below are conducted on selected test-set trajectories constructed from these yearly records.

The data are preprocessed using a LogMinMax normalization procedure. First, a logarithmic transformation is applied to reduce the heavy right-skew in the energy-consumption distribution:
\begin{equation}
    x'=\log(1+x).
    \label{eq:log_transform}
\end{equation}
The transformed data are then scaled to $[0,1]$ using statistics computed from the training set. Inverse transformations are applied before evaluating forecasting errors, so the reported MAE, MAPE, and RMSE values are computed on the original data scale. Missing values are masked during both training and evaluation.

The pretrained models used in the experiments correspond to different historical sequence lengths. The $1$-to-$1$ model uses one previous time step to predict the next state, the $4$-to-$1$ model uses four previous time steps, and the $12$-to-$1$ model uses twelve previous time steps. All model configurations use one-step forecasting horizon $H=1$.

For the reported figures, we use finite test trajectory segments rather than the entire yearly record. In Example~\ref{sec:example1}, the direct prediction comparison is displayed over $t=0,\ldots,800$. In the data-assimilation experiments, the filtering horizon is $850$ steps, although some figures display a shorter interval for visualization.

\begin{table}[ht]
\caption{Experimental configurations for the pretrained STLLM models. The train/validation/test counts refer to supervised forecasting samples constructed from the hourly records.}
\label{tab:stllm_experiment_configs}
\centering
\small
\begin{tabular}{lcccc}
\toprule
\textbf{Model configuration} & \textbf{Seq. Len.} & \textbf{Batch Size} & \textbf{Train/Val/Test} & \textbf{Epochs} \\
\midrule
$1$-to-$1$ model & 1  & 16 & 6980 / 872 / 872 & 289 \\
$12$-to-$1$ model & 12 & 1  & 6980 / 872 / 872 & 126 \\
$4$-to-$1$ model, full training split & 4  & 4  & 6980 / 872 / 872 & 149 \\
$4$-to-$1$ model, reduced training split & 4  & 4  & 1746 / 872 / 872 & 113 \\
\bottomrule
\end{tabular}
\end{table}

In the numerical examples below, the $1$-to-$1$, $4$-to-$1$, and $12$-to-$1$ models refer to pretrained STLLM models with sequence lengths $1$, $4$, and $12$, respectively. The $4$-to-$1$ model trained with the reduced training split is used to represent an insufficiently trained forward model, while the $4$-to-$1$ model trained with the full training split is used to represent a sufficiently trained forward model.

\begin{table}[ht]
\caption{STLLM model hyperparameters.}
\label{tab:stllm_hyperparams}
\centering
\small
\begin{tabular}{lc}
\toprule
\textbf{Hyperparameter} & \textbf{Value} \\
\midrule
Model dimension $d$ & 64 \\
Number of attention heads $h$ & 8 \\
Head dimension $d_h$ & 8 \\
Feed-forward dimension $d_{ff}$ & 384 \\
Number of STLLM blocks $L$ & 4 \\
Dropout rate & 0.1 \\
Total parameters & 749{,}057 \\
\bottomrule
\end{tabular}
\end{table}

The models are trained using AdamW with initial learning rate $10^{-3}$ and weight decay $10^{-4}$. A cosine annealing schedule decreases the learning rate from $10^{-3}$ to $10^{-6}$ over a maximum of $2{,}000$ epochs. The training loss is the masked mean absolute error over non-missing entries:
\begin{equation}
    \mathcal{L}
    =
    \frac{1}{|\mathcal{M}|}
    \sum_{(i,j,k)\in \mathcal{M}}
    \left|
        \widehat{y}_{i,j,k}-y_{i,j,k}
    \right|,
    \label{eq:masked_mae_loss}
\end{equation}
where $\mathcal{M}$ denotes the set of observed non-NaN entries. Gradient norms are clipped to a maximum value of $5$, and early stopping is applied when the validation MAE does not improve for $30$ consecutive epochs.

\FloatBarrier

\subsection{Data assimilation setup}
\label{sec:numerical_setup}

In the numerical experiments, the state vector represents the energy-consumption values of $5000$ users. Thus, we set
\[
    X_{t_n}\in\mathbb{R}^{d_x},\qquad d_x=5000.
\]
The real recorded test-set trajectory is treated as the reference trajectory and is denoted by $X^{\mathrm{true}}_{t_n}$. The pretrained forecasting model introduced above is used as the black-box forward propagator.

For the data-assimilation experiments based on the $4$-to-$1$ model, the input window $W_n$ contains four consecutive state vectors. The initial input window is taken from the reference trajectory. After assimilation begins, the input window is updated recursively by data-assimilation-corrected states: at each filtering step, the oldest state in the four-step window is removed and the newly corrected state is appended. Therefore, after initialization, the learned forward model is driven by assimilated states rather than by the true trajectory.

All filtering computations are performed on the LogMinMax-normalized scale used by the pretrained forecasting model. The observational noise is also added on this normalized scale. Specifically, for the observation model, we use Gaussian noise
$$
    \epsilon_{t_{n+1}}\sim \mathcal{N}(0,\sigma_{\rm obs}^2 I),
    \qquad
    \sigma_{\rm obs}=0.05.
$$
After filtering, the estimated states are transformed back to the original energy-consumption scale before computing the error metrics. Unless otherwise stated, the data assimilation experiments are run for $850$ filtering steps.

For the Ensemble Score Filter, we use an ensemble size $M=50$ and discretize the pseudo-time interval $[0,1]$ in the reverse diffusion process using $L=500$ steps. The damping function in the posterior score update is chosen as
$$
    g(\tau)=1-\tau.
$$
For the EnKF comparison in Example~\ref{sec:example3}, we use the same pretrained forward model, ensemble size, observation masks, observation noise level, and normalized data scale as in the EnSF experiment.

Partial observations are implemented through blockwise observation masks. For an observation percentage $100/B$, the state vector is partitioned into $B$ contiguous blocks with approximately equal sizes. At filtering step $n$, the block indexed by $n \mod B$ is observed. Thus, for $25\%$ observation, one quarter of the state components is observed at each filtering step, and every component is observed once every four steps. In the main numerical examples, we report results for observation levels $25\%$, $50\%$, and $100\%$.

In Example~\ref{sec:example2}, we use direct partial observations. The observation model is
\begin{equation}
    Y_{t_{n+1}}
    =
    \operatorname{Mask}
    \left(
        X_{t_{n+1}}+\epsilon_{t_{n+1}}
    \right),
    \label{eq:numerical_direct_obs}
\end{equation}
where $\operatorname{Mask}(\cdot)$ selects the observed block at the current filtering step. In Example~\ref{sec:example3}, we compare EnSF with EnKF under a mixed observation setting. Among the observed components, half are observed directly and half are observed through the nonlinear arctangent map. Equivalently, for an observed component $i$, the mixed observation operator is given by
\begin{equation}
    \mathcal{H}_{mix}(X_i)
    =
    \begin{cases}
        X_i, & \text{for directly observed components},\\
        \arctan(X_i), & \text{for nonlinear observed components}.
    \end{cases}
    \label{eq:numerical_mixed_obs}
\end{equation}
The corresponding observation is then
\begin{equation}
    Y_{t_{n+1}}
    =
    \operatorname{Mask}
    \left(
        \mathcal{H}_{mix}(X_{t_{n+1}})
        +
        \epsilon_{t_{n+1}}
    \right).
    \label{eq:numerical_mixed_obs_full}
\end{equation}

We evaluate the forecasting and filtering results using mean absolute error (MAE), mean absolute percentage error (MAPE), and root mean squared error (RMSE). For a prediction or filtering estimate $\tilde{X}_{t_n}$ and reference state $X^{\mathrm{true}}_{t_n}$, these metrics are defined by
\begin{equation}
    \operatorname{MAE}(t_n)
    =
    \frac{1}{d_x}
    \sum_{i=1}^{d_x}
    \left|
        \tilde{X}_{t_n}^{i}
        -
        \left(X^{\mathrm{true}}_{t_n}\right)^{i}
    \right|,
    \label{eq:mae_metric}
\end{equation}
\begin{equation}
    \operatorname{MAPE}(t_n)
    =
    \frac{100}{d_x}
    \sum_{i=1}^{d_x}
    \left|
        \frac{
            \tilde{X}_{t_n}^{i}
            -
            \left(X^{\mathrm{true}}_{t_n}\right)^{i}
        }{
            \left(X^{\mathrm{true}}_{t_n}\right)^{i}
        }
    \right|,
    \label{eq:mape_metric}
\end{equation}
and
\begin{equation}
    \operatorname{RMSE}(t_n)
    =
    \sqrt{
    \frac{1}{d_x}
    \sum_{i=1}^{d_x}
    \left(
        \tilde{X}_{t_n}^{i}
        -
        \left(X^{\mathrm{true}}_{t_n}\right)^{i}
    \right)^2
    }.
    \label{eq:rmse_metric}
\end{equation}
All three metrics are computed after inverse normalization, so the reported errors are measured in the original energy-consumption scale. In Examples~\ref{sec:example2} and \ref{sec:example3}, the RMSE curves are computed over all $5000$ state components. The trajectory plots, however, are shown only for selected representative dimensions in order to visualize the temporal behavior of individual users.

\subsection{Example 1. Model comparison for the $1$-to-$1$, $4$-to-$1$, and $12$-to-$1$ forward models}
\label{sec:example1}

In this example, we compare the forecasting performance of the $1$-to-$1$, $4$-to-$1$, and $12$-to-$1$ pretrained forward models on real energy-consumption data. We first consider the setting in which the model is supplied with correct input information at each prediction step. We then consider an open-loop setting in which the model output is recursively used as input without corrective information. The purpose of this example is to evaluate the baseline predictive capability of the pretrained models and to demonstrate why data assimilation is needed for long-horizon forecast correction.

We first compare the direct prediction performance of the three models over the time interval $t=0,\dots,800$. For each model, we compare the model prediction with the corresponding real result. The error metrics are computed according to the definitions in Section~\ref{sec:numerical_setup}.

\begin{figure}[ht]
    \centering
    \begin{subfigure}[t]{0.32\textwidth}
        \centering
        \includegraphics[width=\textwidth]{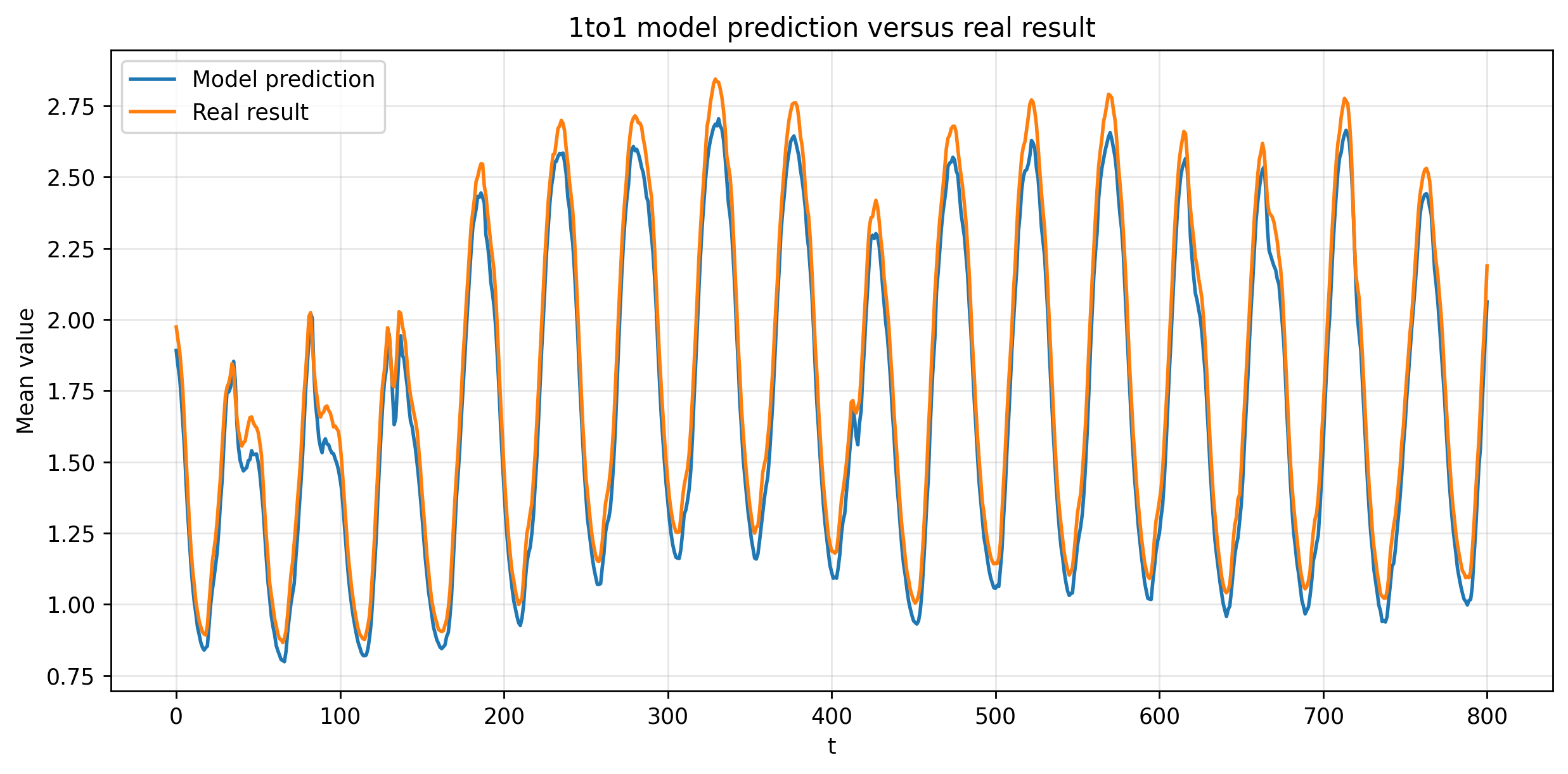}
        \caption{$1$-to-$1$ model prediction versus real result.}
        \label{fig:example1_1to1}
    \end{subfigure}
    \hfill
    \begin{subfigure}[t]{0.32\textwidth}
        \centering
        \includegraphics[width=\textwidth]{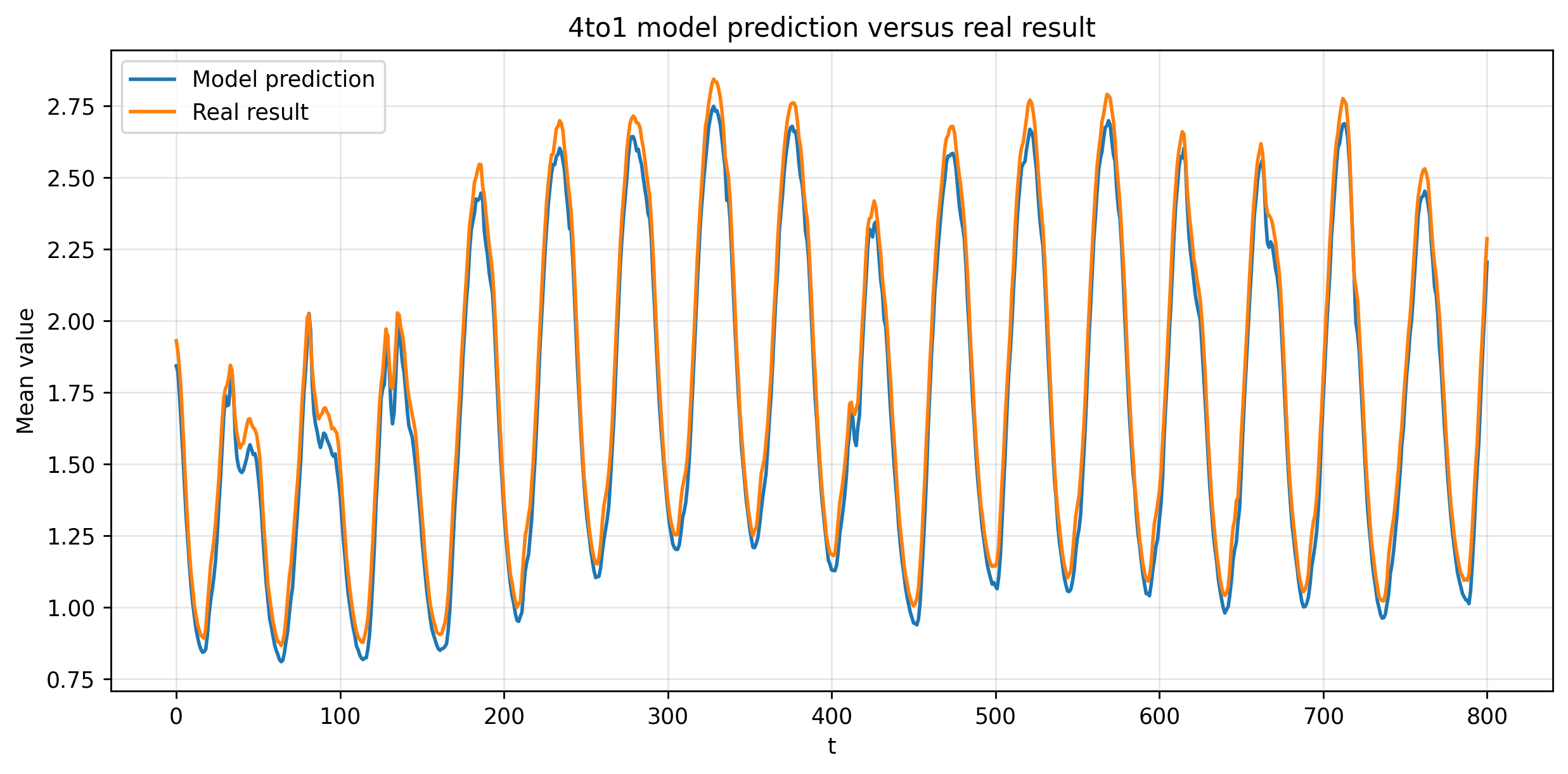}
        \caption{$4$-to-$1$ model prediction versus real result.}
        \label{fig:example1_4to1}
    \end{subfigure}
    \hfill
    \begin{subfigure}[t]{0.32\textwidth}
        \centering
        \includegraphics[width=\textwidth]{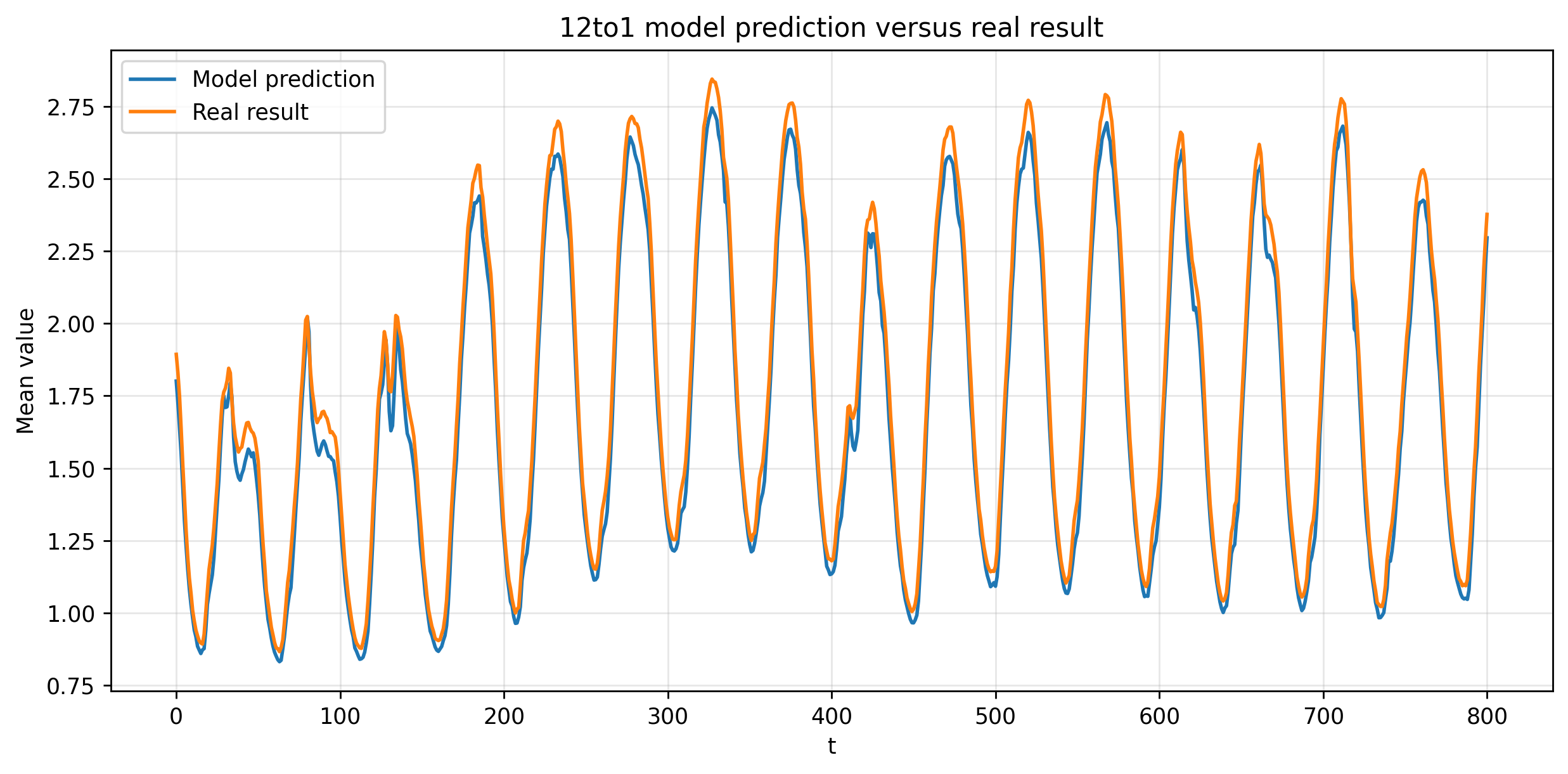}
        \caption{$12$-to-$1$ model prediction versus real result.}
        \label{fig:example1_12to1}
    \end{subfigure}
    \caption{Comparison between model prediction and real result for the $1$-to-$1$, $4$-to-$1$, and $12$-to-$1$ forward models over $t=0,\dots,800$, when correct input information is supplied at each prediction step.}
    \label{fig:example1_pred_vs_real}
\end{figure}

Fig.~\ref{fig:example1_pred_vs_real} shows the prediction curves and the corresponding real results for the three models. In all three cases, the learned forward model captures the dominant oscillatory pattern of the underlying energy-consumption dynamics. At the same time, visible discrepancies remain between the prediction and the real result, especially near peaks and troughs. Nevertheless, the one-step prediction results remain reasonably accurate, and the longer-input models exhibit better agreement with the real signal.

\begin{figure}[ht]
    \centering
    \begin{subfigure}[t]{0.32\textwidth}
        \centering
        \includegraphics[width=\textwidth]{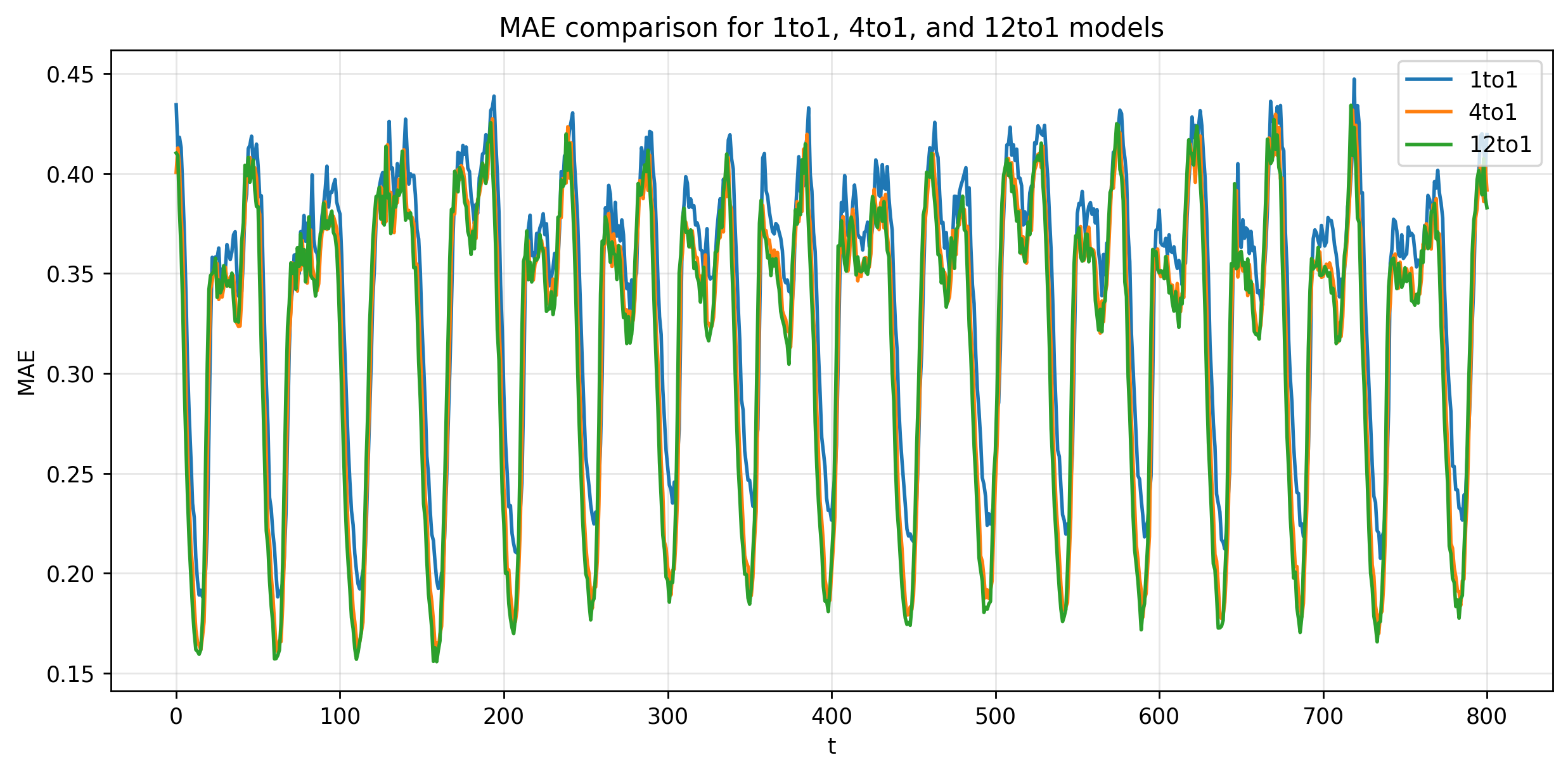}
        \caption{MAE comparison.}
        \label{fig:example1_mae}
    \end{subfigure}
    \hfill
    \begin{subfigure}[t]{0.32\textwidth}
        \centering
        \includegraphics[width=\textwidth]{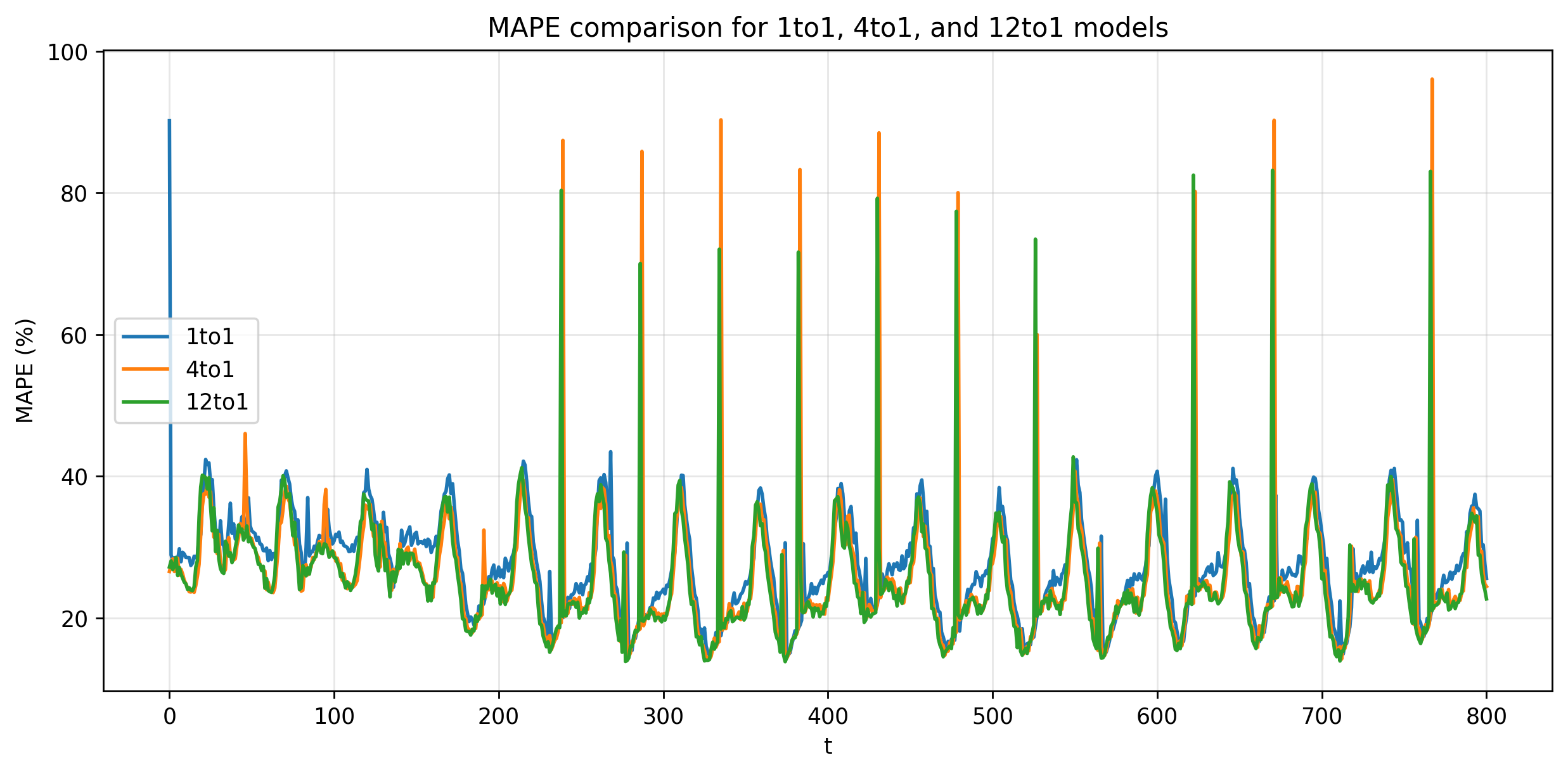}
        \caption{MAPE comparison.}
        \label{fig:example1_mape}
    \end{subfigure}
    \hfill
    \begin{subfigure}[t]{0.32\textwidth}
        \centering
        \includegraphics[width=\textwidth]{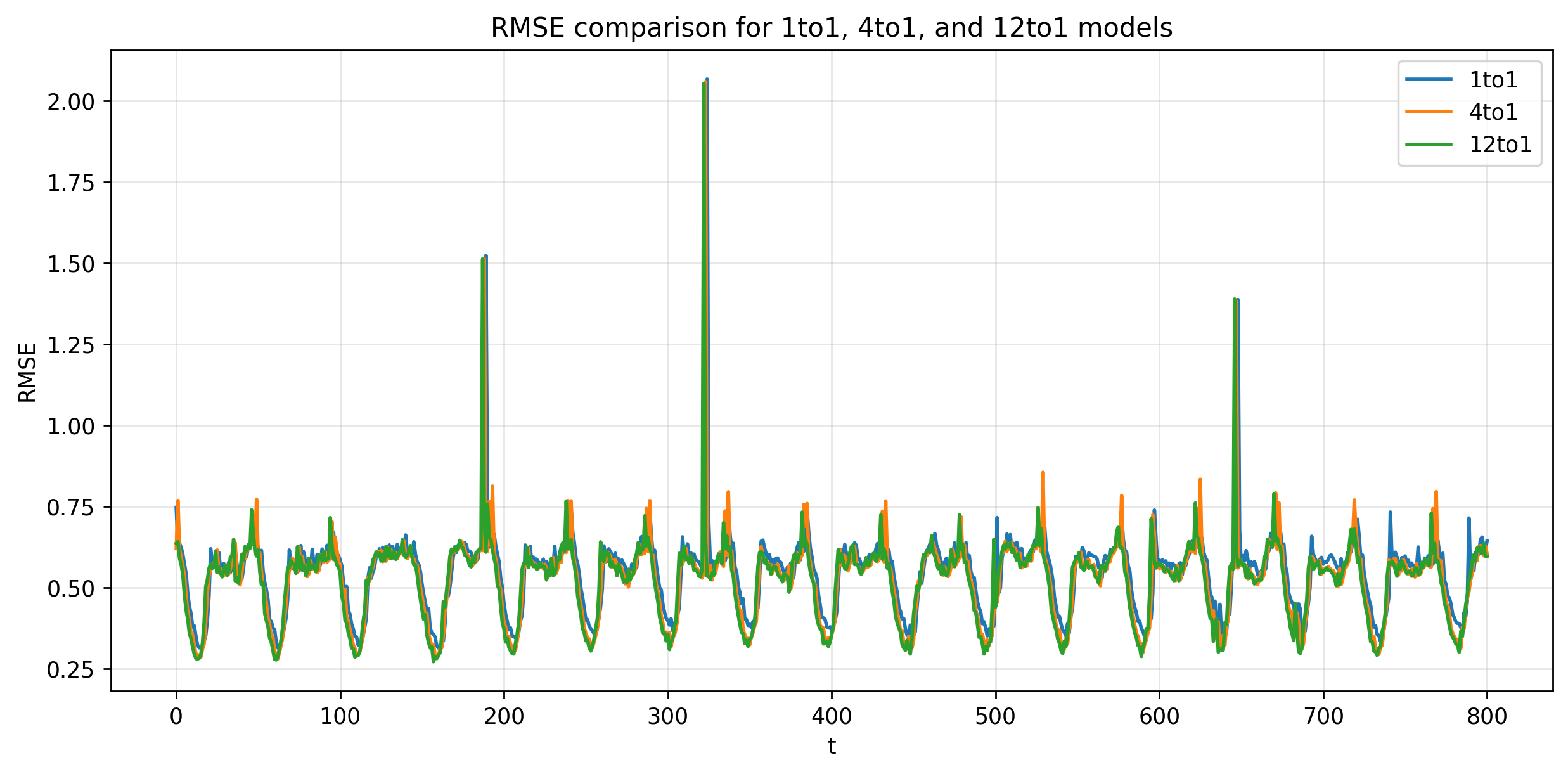}
        \caption{RMSE comparison.}
        \label{fig:example1_rmse}
    \end{subfigure}
    \caption{Comparison of time-dependent MAE, MAPE, and RMSE for the $1$-to-$1$, $4$-to-$1$, and $12$-to-$1$ forward models over $t=0,\dots,800$, when correct input information is supplied at each prediction step.}
    \label{fig:example1_metrics}
\end{figure}

To provide a more direct comparison among the three models, we plot the time-dependent MAE, MAPE, and RMSE in Fig.~\ref{fig:example1_metrics}. The three models exhibit similar temporal error patterns, indicating that they are driven by the same underlying demand dynamics. However, the $4$-to-$1$ and $12$-to-$1$ models generally achieve lower errors than the $1$-to-$1$ model, and the $12$-to-$1$ model gives the best overall performance among the three. The MAPE curves contain several sharp spikes, which are expected when the true values become relatively small, since percentage-based errors are more sensitive in that regime.

For completeness, we also report the average error statistics over the whole time interval in Table~\ref{tab:example1_metrics}. The results show that both the $4$-to-$1$ and $12$-to-$1$ models improve upon the $1$-to-$1$ model in all three metrics. Among the three models, the $12$-to-$1$ model attains the lowest average MAE, MAPE, and RMSE, indicating the best overall predictive accuracy in this direct-prediction setting.

\begin{table}[ht]
\caption{Average error statistics for the $1$-to-$1$, $4$-to-$1$, and $12$-to-$1$ models over $t=0,\dots,800$ when correct input information is supplied at each prediction step. Lower values indicate better performance.}
\label{tab:example1_metrics}
\centering
\small
\begin{tabular}{lccc}
\toprule
\textbf{Model} & \textbf{MAE} & \textbf{MAPE (\%)} & \textbf{RMSE} \\
\midrule
$1$-to-$1$   & 0.340168 & 27.359060 & 0.544696 \\
$4$-to-$1$   & 0.319658 & 25.876973 & 0.526702 \\
$12$-to-$1$  & 0.317984 & 25.661881 & 0.522809 \\
\bottomrule
\end{tabular}
\end{table}

We next consider the same three pretrained forward models in an open-loop setting. In this case, after the initial input is provided, the forecast is propagated solely by the learned model itself, without using corrective information from the true trajectory. This setting is more difficult because forecast errors are no longer corrected at each step and may accumulate over time. The purpose of this part of the example is to show that a pretrained forward model alone is not sufficient for reliable long-horizon state tracking.

\begin{figure}[ht]
    \centering
    \begin{subfigure}[t]{0.32\textwidth}
        \centering
        \includegraphics[width=\textwidth]{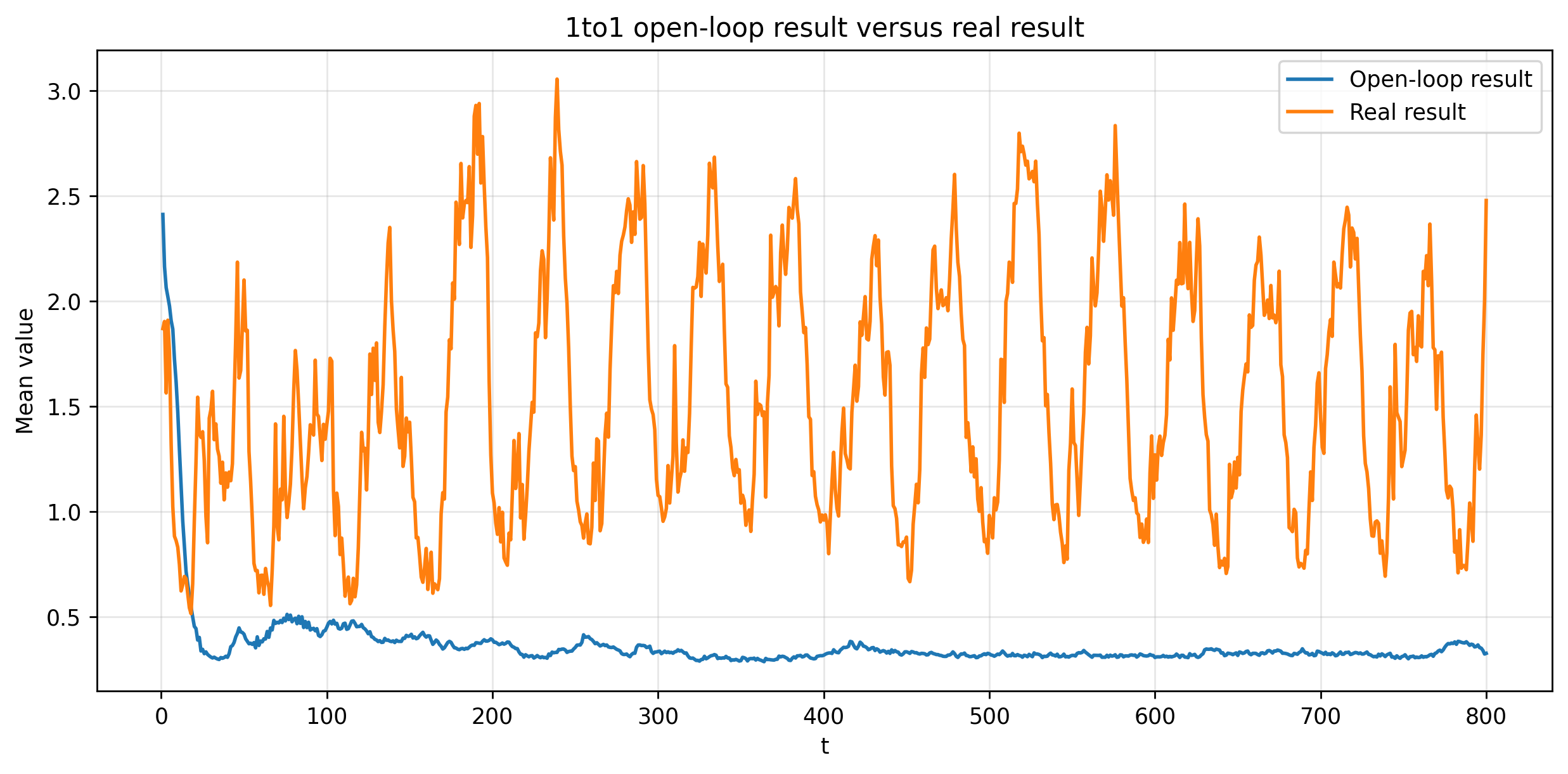}
        \caption{$1$-to-$1$ open-loop result versus real result.}
        \label{fig:example1_openloop_1to1}
    \end{subfigure}
    \hfill
    \begin{subfigure}[t]{0.32\textwidth}
        \centering
        \includegraphics[width=\textwidth]{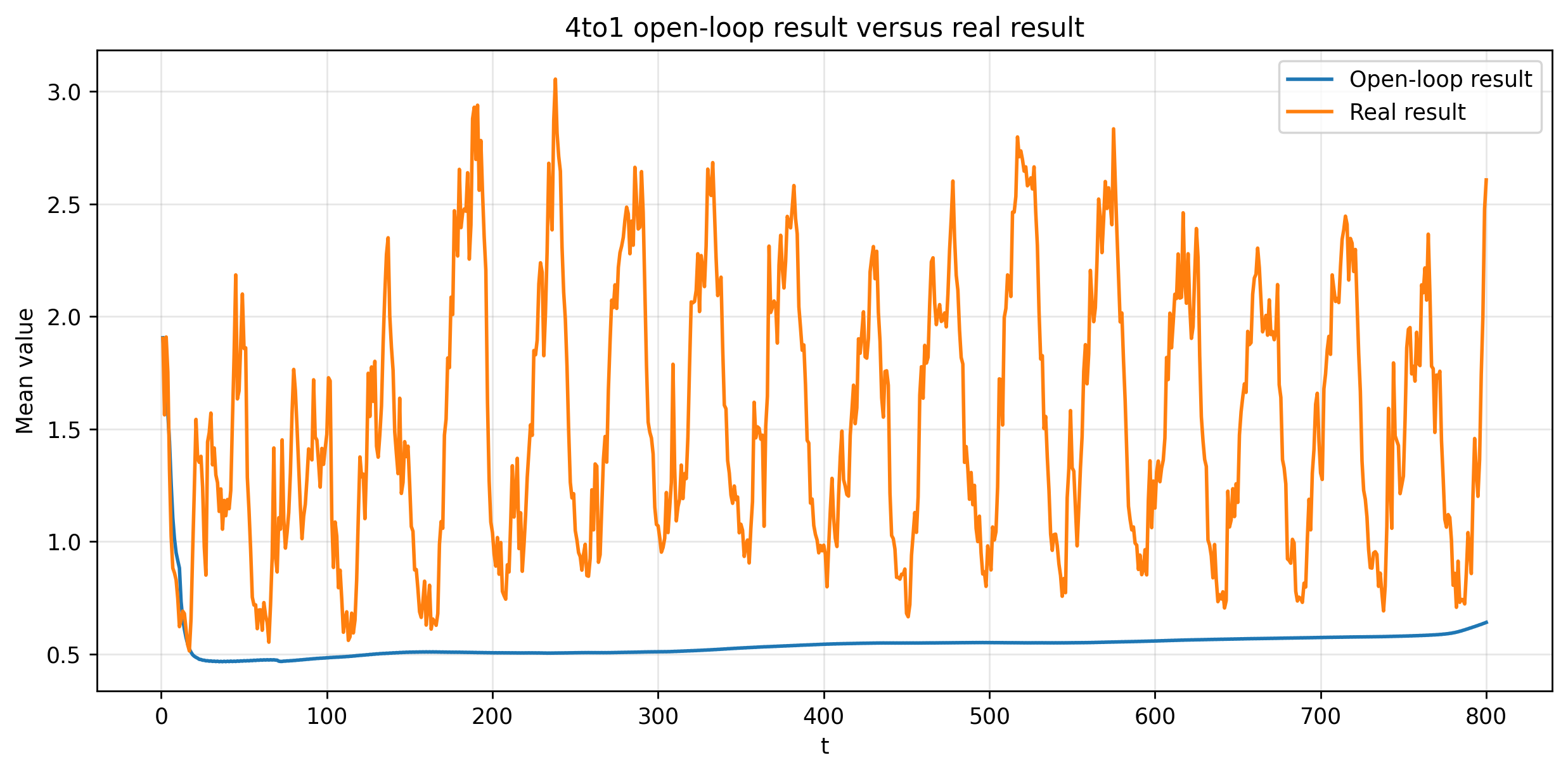}
        \caption{$4$-to-$1$ open-loop result versus real result.}
        \label{fig:example1_openloop_4to1}
    \end{subfigure}
    \hfill
    \begin{subfigure}[t]{0.32\textwidth}
        \centering
        \includegraphics[width=\textwidth]{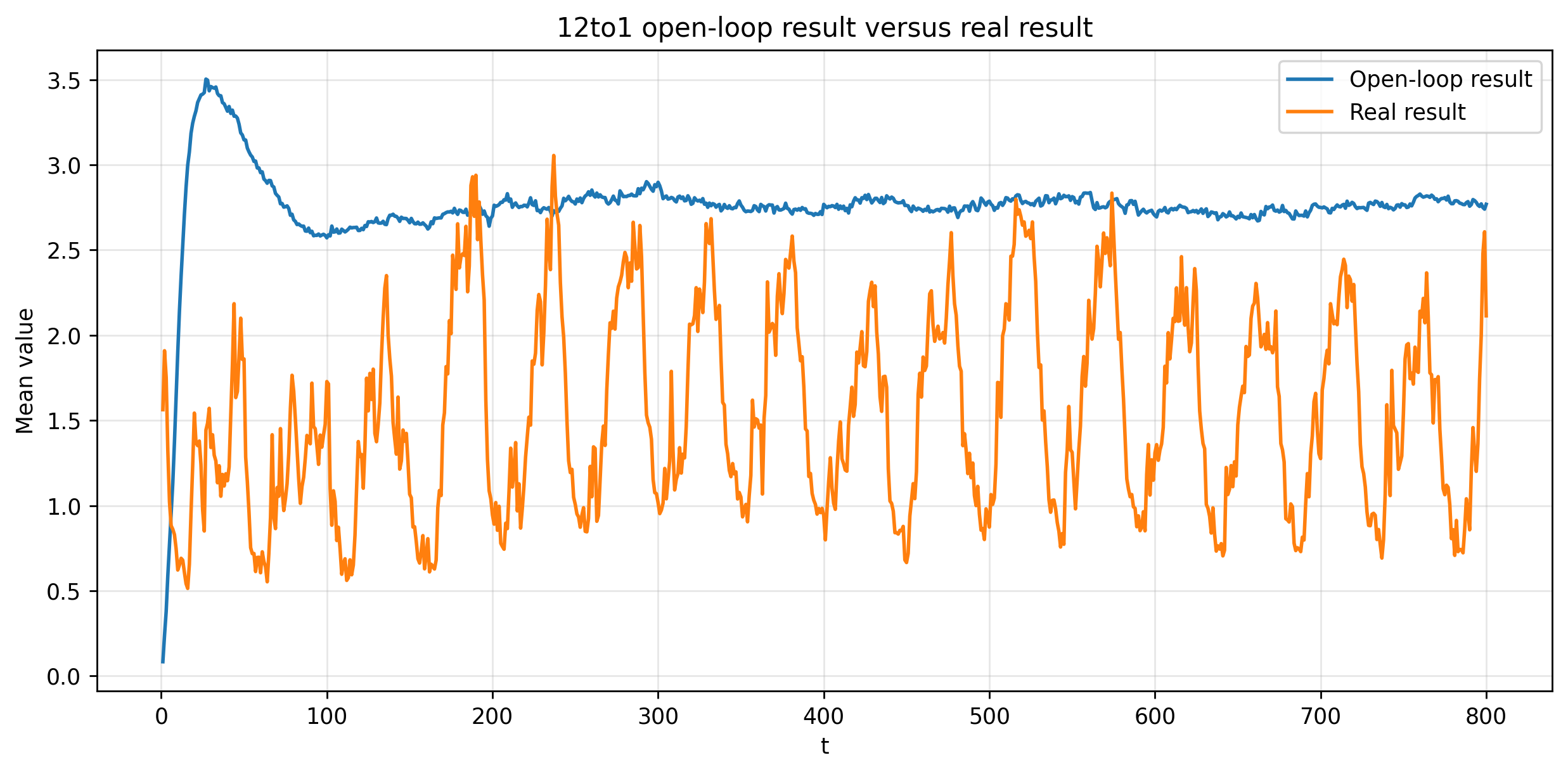}
        \caption{$12$-to-$1$ open-loop result versus real result.}
        \label{fig:example1_openloop_12to1}
    \end{subfigure}
    \caption{Comparison between open-loop forecasts and real results for the $1$-to-$1$, $4$-to-$1$, and $12$-to-$1$ forward models. The forecast is propagated without corrective information. The comparison is shown on the saved subset of representative state entries.}
    \label{fig:example1_openloop_vs_real}
\end{figure}

The open-loop results are compared with the real results in Fig.~\ref{fig:example1_openloop_vs_real}. Here, the comparison is performed on a saved subset of representative state entries rather than on the full state vector, since the open-loop diagnostic data were stored only for those entries. Even with this reduced comparison, the main behavior is clear: without sequential correction, the open-loop forecasts lose agreement with the real trajectory. The discrepancy is already visible for the $1$-to-$1$ and $4$-to-$1$ models and is especially severe for the $12$-to-$1$ model.

\begin{figure}[ht]
    \centering
    \begin{subfigure}[t]{0.32\textwidth}
        \centering
        \includegraphics[width=\textwidth]{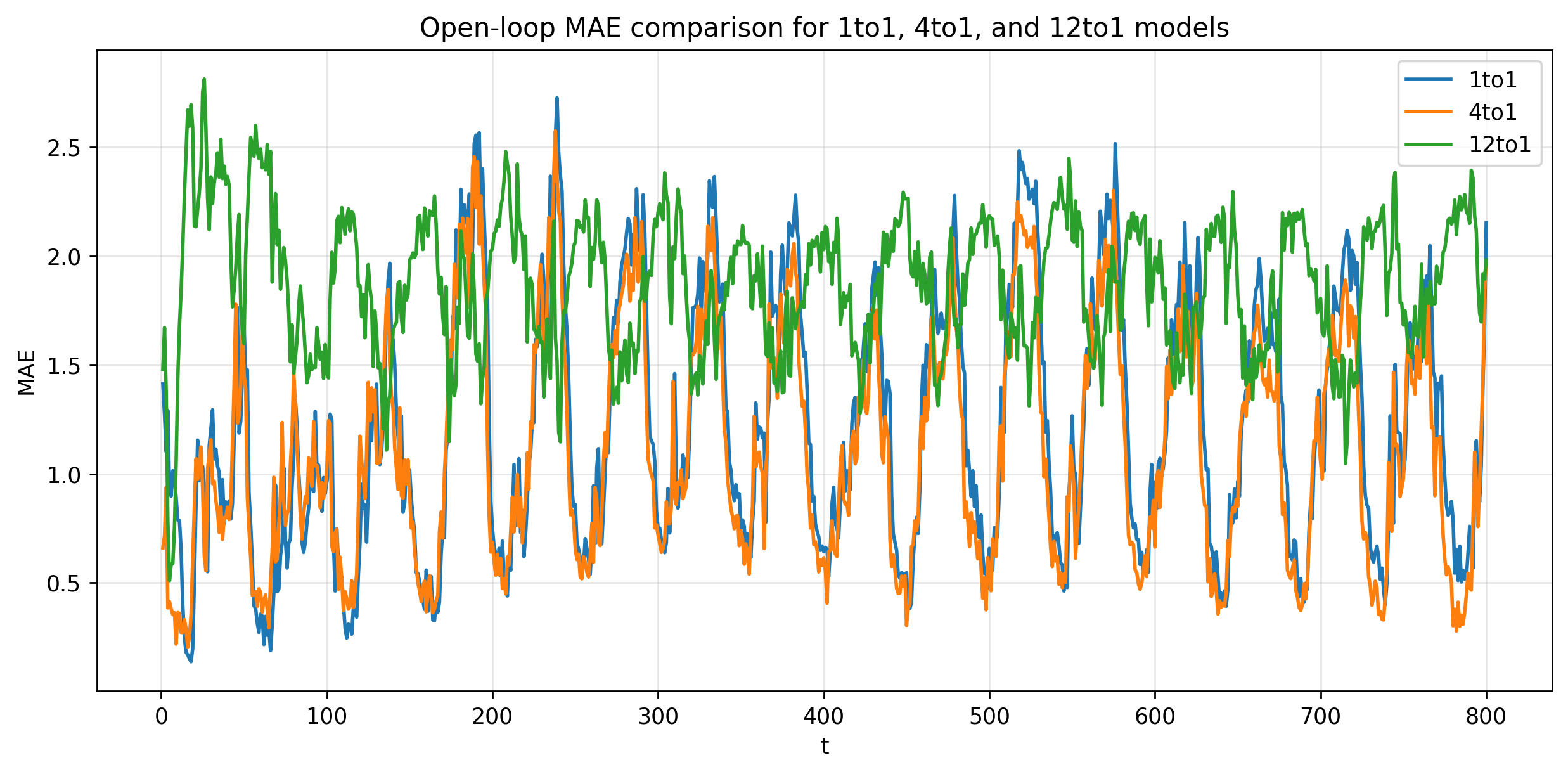}
        \caption{MAE comparison.}
        \label{fig:example1_openloop_mae}
    \end{subfigure}
    \hfill
    \begin{subfigure}[t]{0.32\textwidth}
        \centering
        \includegraphics[width=\textwidth]{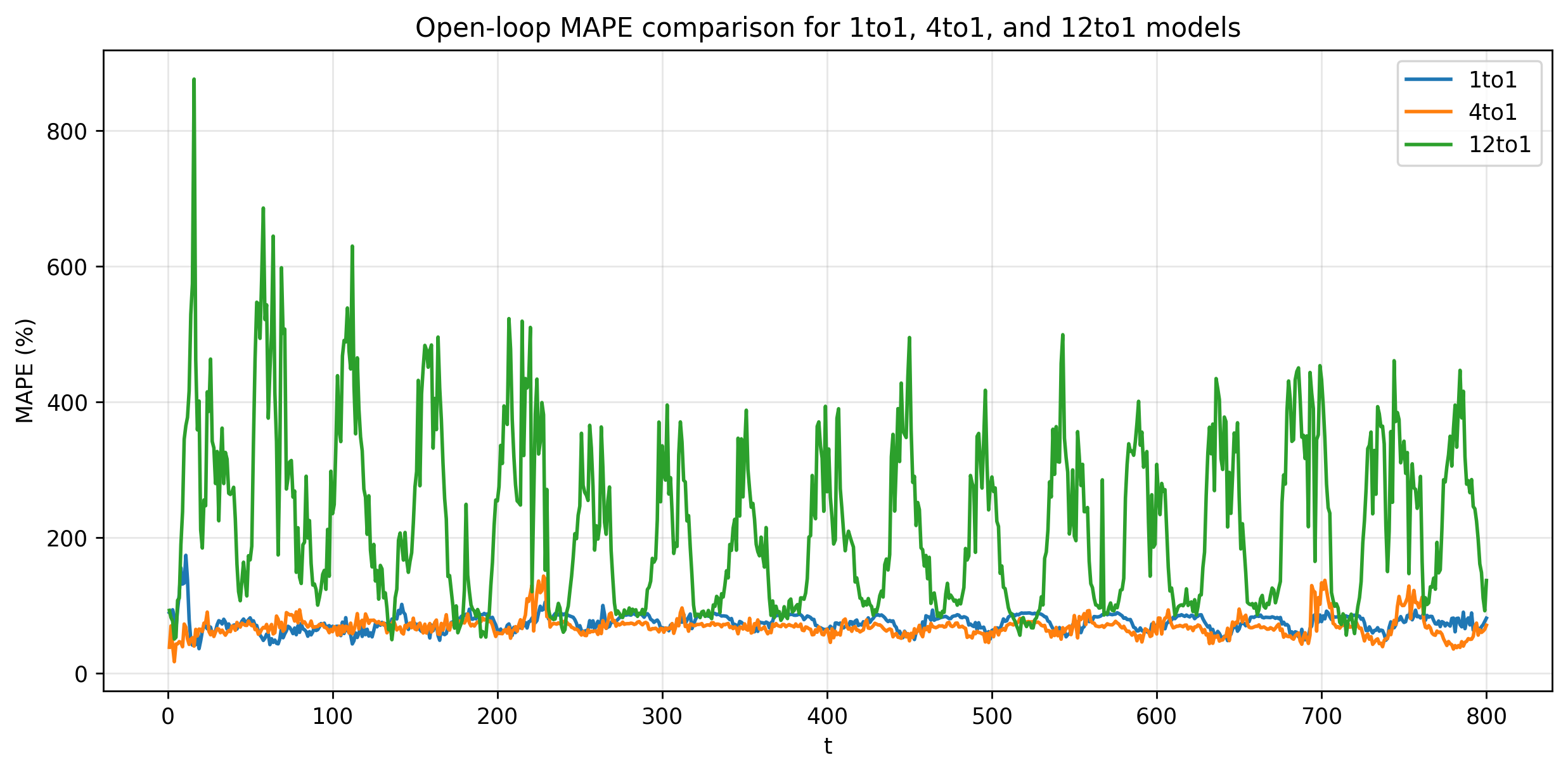}
        \caption{MAPE comparison.}
        \label{fig:example1_openloop_mape}
    \end{subfigure}
    \hfill
    \begin{subfigure}[t]{0.32\textwidth}
        \centering
        \includegraphics[width=\textwidth]{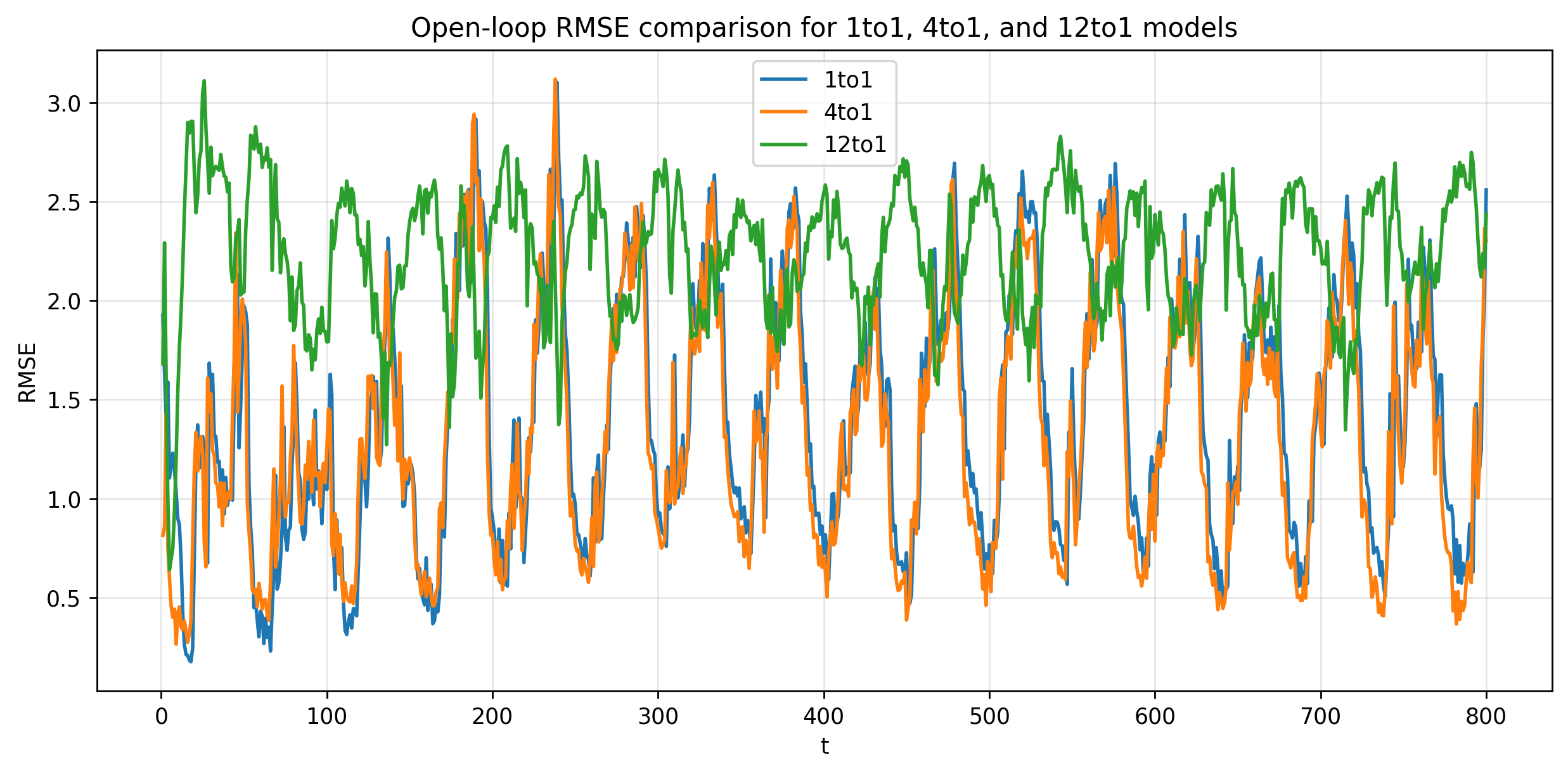}
        \caption{RMSE comparison.}
        \label{fig:example1_openloop_rmse}
    \end{subfigure}
    \caption{Comparison of time-dependent MAE, MAPE, and RMSE for the open-loop forecasts of the $1$-to-$1$, $4$-to-$1$, and $12$-to-$1$ models. The metrics are computed on the saved subset of representative state entries.}
    \label{fig:example1_openloop_metrics}
\end{figure}

The time-dependent MAE, MAPE, and RMSE for the open-loop trajectories are shown in Fig.~\ref{fig:example1_openloop_metrics}. The error levels are much larger than those in the direct-prediction comparison. Among the three models, the $4$-to-$1$ model gives the smallest average open-loop errors, while the $12$-to-$1$ model performs the worst by a large margin. This is particularly notable because the $12$-to-$1$ model performed best in the direct-prediction setting. Therefore, strong short-horizon predictive accuracy does not by itself guarantee reliable long-horizon open-loop performance.

The average open-loop error statistics are reported in Table~\ref{tab:example1_openloop_metrics}. These values confirm the qualitative conclusions drawn from the plots. In particular, all three models exhibit substantial degradation relative to the direct-prediction setting, and the deterioration is especially large for the $12$-to-$1$ model. This part of the example provides the main motivation for the data-assimilation results that follow: although the pretrained forward models can achieve reasonable short-horizon prediction accuracy, their open-loop forecasts can become unreliable when corrective information is no longer supplied. Sequential data assimilation is therefore needed to incorporate observational information and improve state estimation over time.

\begin{table}[ht]
\caption{Average error statistics for the open-loop forecasts of the $1$-to-$1$, $4$-to-$1$, and $12$-to-$1$ models. The metrics are computed on the saved subset of representative state entries. Lower values indicate better performance.}
\label{tab:example1_openloop_metrics}
\centering
\small
\begin{tabular}{lccc}
\toprule
\textbf{Model} & \textbf{MAE} & \textbf{MAPE (\%)} & \textbf{RMSE} \\
\midrule
$1$-to-$1$   & 1.211562 & 73.961318  & 1.392503 \\
$4$-to-$1$   & 1.106473 & 68.135904  & 1.319839 \\
$12$-to-$1$  & 1.868828 & 223.743966 & 2.236019 \\
\bottomrule
\end{tabular}
\end{table}

\subsection{Example 2. EnSF correction for $4$-to-$1$ models under different training schemes}
\label{sec:example2}

Example~\ref{sec:example1} shows that the pretrained forward models can provide reasonable short-horizon predictions when correct input information is supplied, but their open-loop forecasts may become unreliable when no corrective information is used. In this example, we apply EnSF-based data assimilation to improve the forecast results for the $4$-to-$1$ model under two training configurations. The first model is trained using the reduced training split and is therefore treated as an insufficiently trained forward model. The second model is trained using the full training split and is treated as a sufficiently trained forward model.

For both cases, the pretrained $4$-to-$1$ model is used as the black-box forward propagator, while EnSF sequentially assimilates direct partial observations as described in \eqref{eq:numerical_direct_obs}. We compare the no-DA open-loop result with EnSF-corrected results under observation levels $25\%$, $50\%$, and $100\%$. The RMSE curves in this example are computed over all $5000$ state components after inverse normalization.

\begin{figure}[ht]
    \centering
    \begin{subfigure}[t]{0.48\textwidth}
        \centering
        \includegraphics[width=\textwidth]{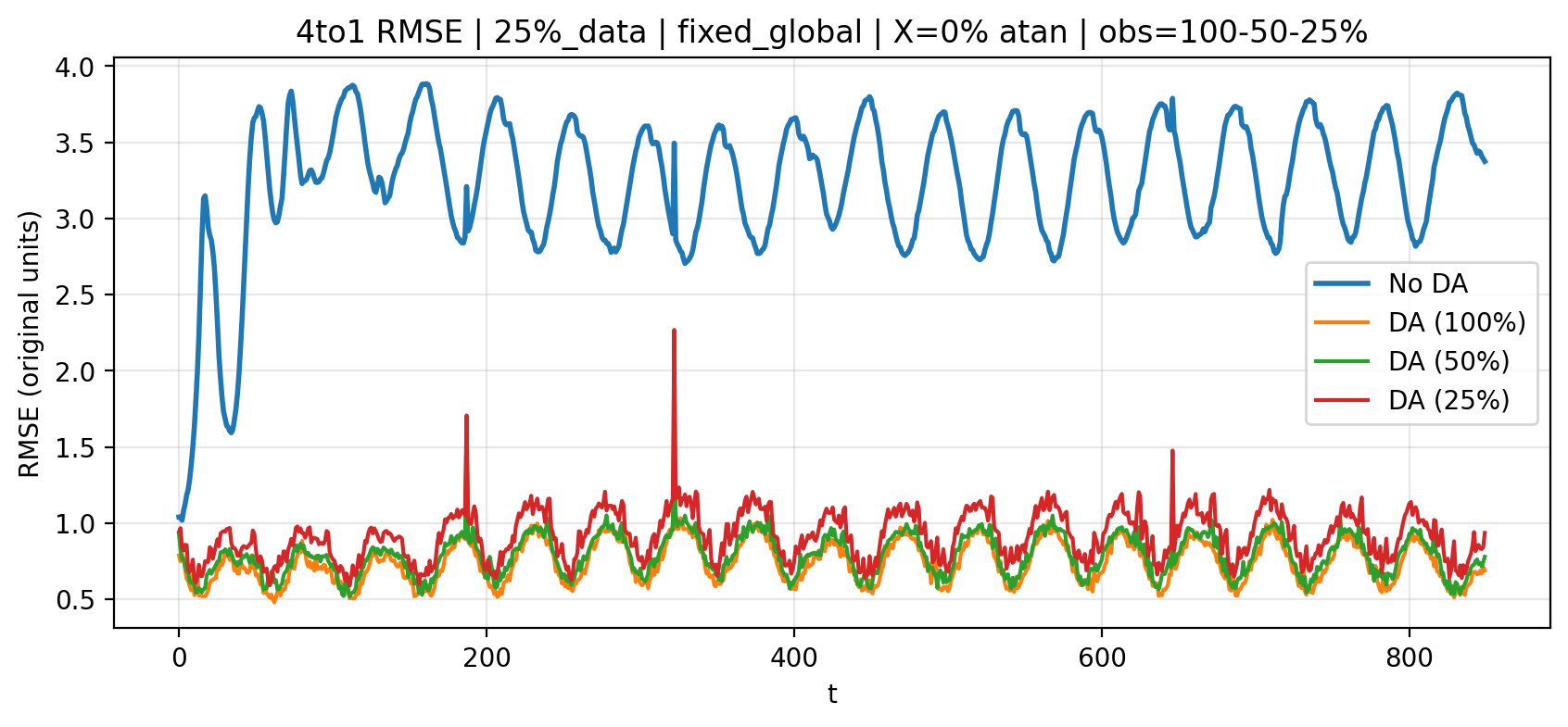}
        \caption{$4$-to-$1$ model trained with the reduced training split.}
        \label{fig:example2_rmse_25}
    \end{subfigure}
    \hfill
    \begin{subfigure}[t]{0.48\textwidth}
        \centering
        \includegraphics[width=\textwidth]{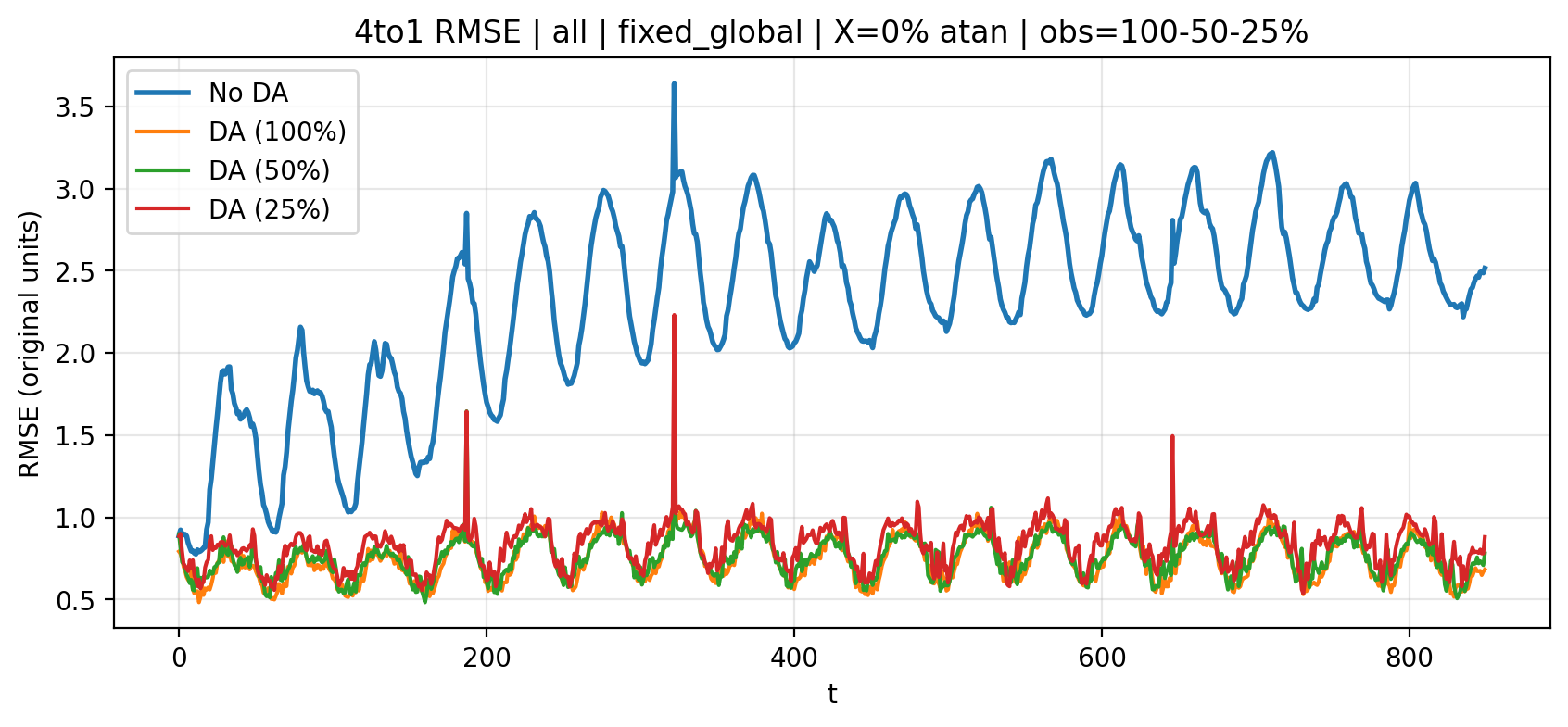}
        \caption{$4$-to-$1$ model trained with the full training split.}
        \label{fig:example2_rmse_all}
    \end{subfigure}
    \caption{RMSE comparison for EnSF correction under different training schemes. The no-DA curve corresponds to the open-loop forecast, while the DA curves correspond to EnSF correction with direct partial observations at $25\%$, $50\%$, and $100\%$ observation levels. The RMSE is computed over all $5000$ state components.}
    \label{fig:example2_rmse}
\end{figure}

Fig.~\ref{fig:example2_rmse} presents the RMSE comparison for the two training schemes. In both cases, the no-DA curve has much larger RMSE than the EnSF-corrected curves, indicating that the learned forward model alone is not sufficient for accurate long-horizon prediction. The improvement is especially clear for the model trained with the reduced training split, where the no-DA forecast error remains large throughout the experiment. After EnSF correction, the RMSE is substantially reduced for all observation levels. As expected, using more observational information generally improves the correction, with the $100\%$ observation case producing the lowest RMSE among the DA results.

\begin{figure}[p]
    \centering
    \begin{subfigure}[t]{0.49\textwidth}
        \centering
        \includegraphics[width=\textwidth]{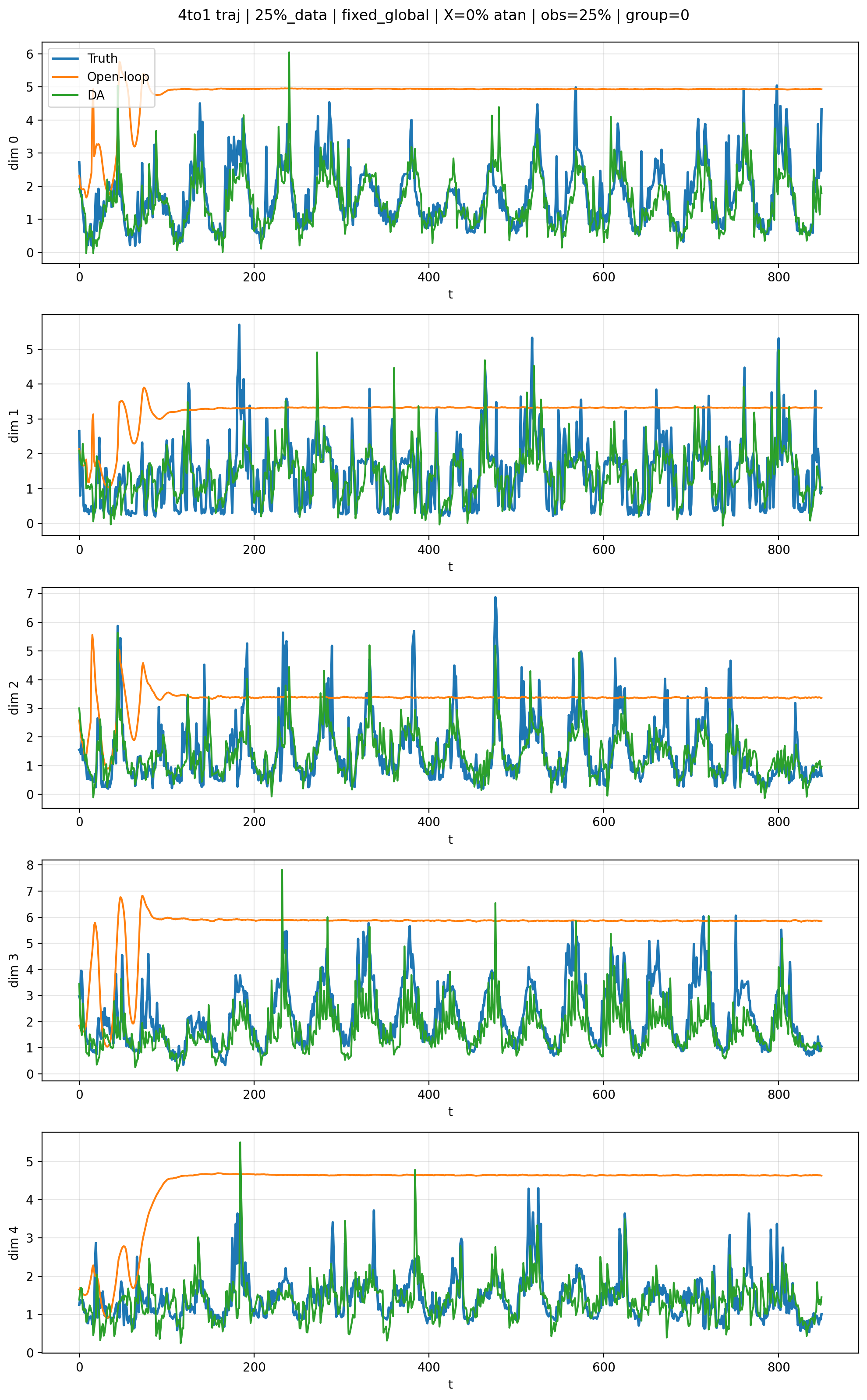}
        \caption{$4$-to-$1$ model trained with the reduced training split, using $25\%$ direct observations.}
        \label{fig:example2_traj_25}
    \end{subfigure}
    \hfill
    \begin{subfigure}[t]{0.49\textwidth}
        \centering
        \includegraphics[width=\textwidth]{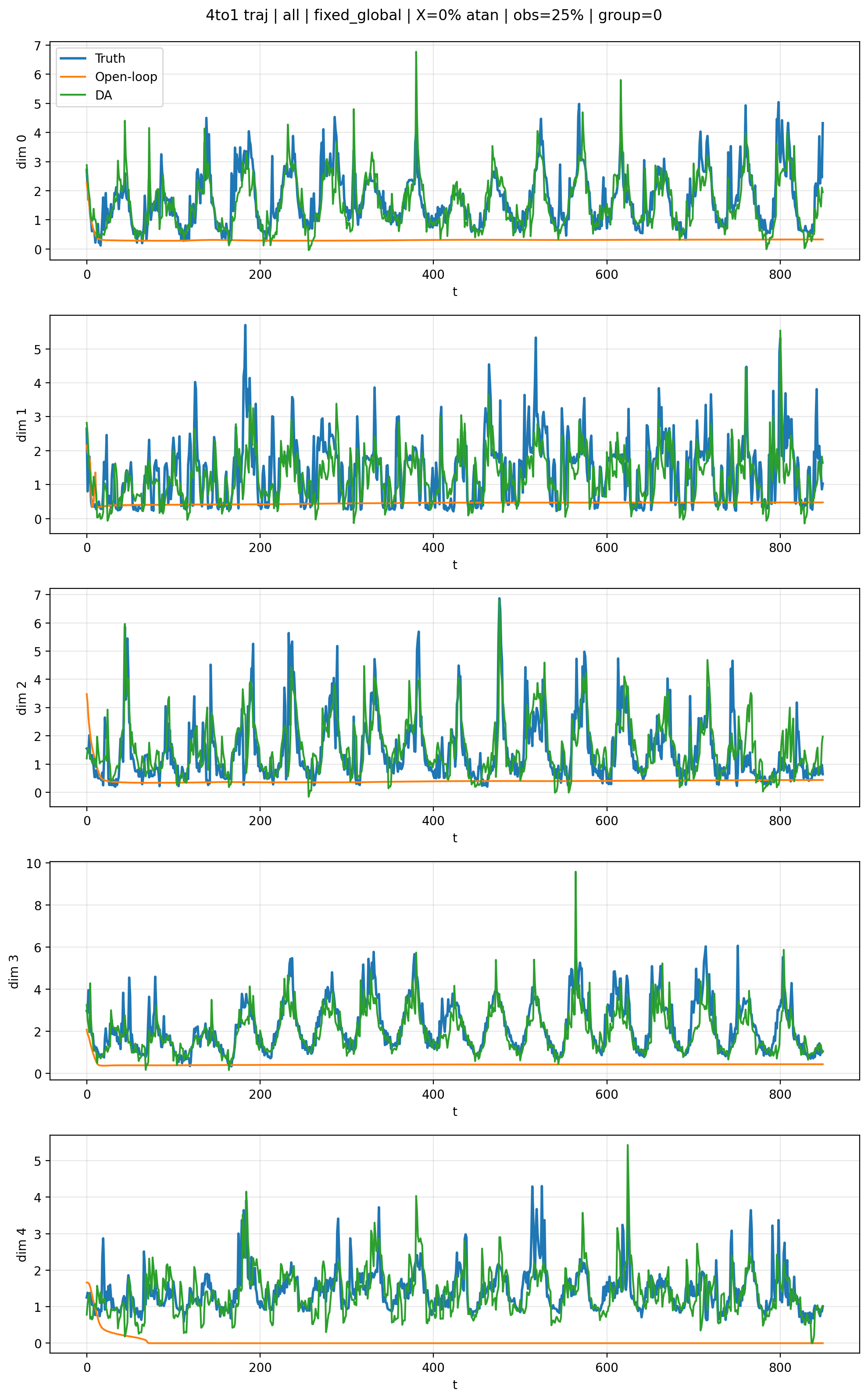}
        \caption{$4$-to-$1$ model trained with the full training split, using $25\%$ direct observations.}
        \label{fig:example2_traj_all}
    \end{subfigure}

    \caption{Representative trajectory comparisons for the $4$-to-$1$ model under EnSF correction. The EnSF-corrected trajectory tracks the truth more closely than the open-loop forecast, even when only partial observations are assimilated. The displayed trajectories correspond to selected representative state dimensions.}
    \label{fig:example2_traj}
\end{figure}

To further illustrate the effect of EnSF at the trajectory level, Fig.~\ref{fig:example2_traj} shows representative trajectory comparisons using $25\%$ observations. These trajectory plots are shown only for selected state dimensions in order to visualize individual temporal behavior. The blue curves denote the truth, the orange curves denote the open-loop forecast, and the green curves denote the EnSF-corrected trajectory. For the insufficiently trained model, the open-loop forecast quickly loses agreement with the true trajectory and remains far from the observed temporal pattern, while the EnSF estimate stays much closer to the truth across the displayed dimensions. For the fully trained model, the open-loop result is different in scale and shape but still fails to reproduce the true temporal variability. In both cases, EnSF provides a clear correction by recovering the main oscillatory structures of the true signal.

These results demonstrate that the benefit of EnSF is not limited to a specific training scheme of the forward model. When the forward model is insufficiently trained, EnSF reduces the forecast error by incorporating observational information into the prediction process. When the forward model is trained on the full training split, EnSF still improves the result by correcting residual inaccuracies in the learned dynamics. Thus, the combined framework of a pretrained spatio-temporal forecasting model and EnSF-based data assimilation provides a more reliable approach for real-data energy-consumption prediction than open-loop model propagation alone.

\FloatBarrier

\subsection{Example 3. Comparison between EnSF and EnKF for the $4$-to-$1$ model}
\label{sec:example3}

In Example~\ref{sec:example2}, we demonstrated that EnSF can substantially improve the forecast results of the $4$-to-$1$ model by assimilating partial observations. We now compare EnSF with the Ensemble Kalman Filter (EnKF) under the same high-dimensional forecast-correction setting. The purpose of this example is to examine whether the score-based update used by EnSF provides an advantage over the Kalman-type ensemble update used by EnKF when the observation operator contains nonlinear components.

We use the $4$-to-$1$ model trained with the reduced training split. This setting is intentionally challenging, since the forward model is not trained on the full training split and therefore provides a less accurate forecast prior. Both EnSF and EnKF use the same pretrained model as the black-box forward propagator and assimilate observations under the same partial-observation masks and noise level. In contrast to Example~\ref{sec:example2}, we use the mixed observation model described in \eqref{eq:numerical_mixed_obs}--\eqref{eq:numerical_mixed_obs_full}, where half of the observed components are observed directly and the other half are observed through the arctangent function. We compare the no-DA open-loop forecast with data-assimilation results using $25\%$, $50\%$, and $100\%$ observation levels.

\begin{figure}[ht]
    \centering
    \begin{subfigure}[t]{0.48\textwidth}
        \centering
        \includegraphics[width=\textwidth]{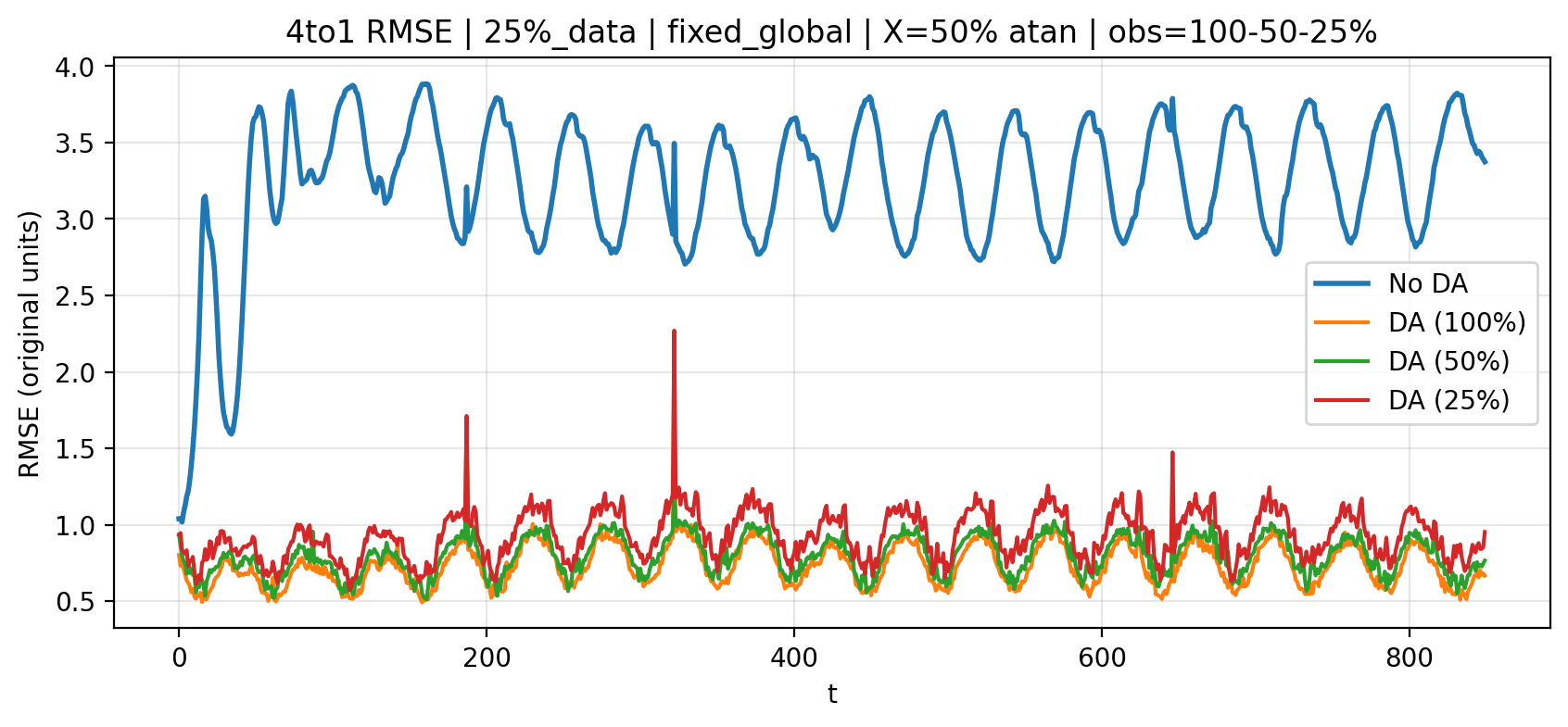}
        \caption{EnSF correction.}
        \label{fig:example3_rmse_ensf}
    \end{subfigure}
    \hfill
    \begin{subfigure}[t]{0.48\textwidth}
        \centering
        \includegraphics[width=\textwidth]{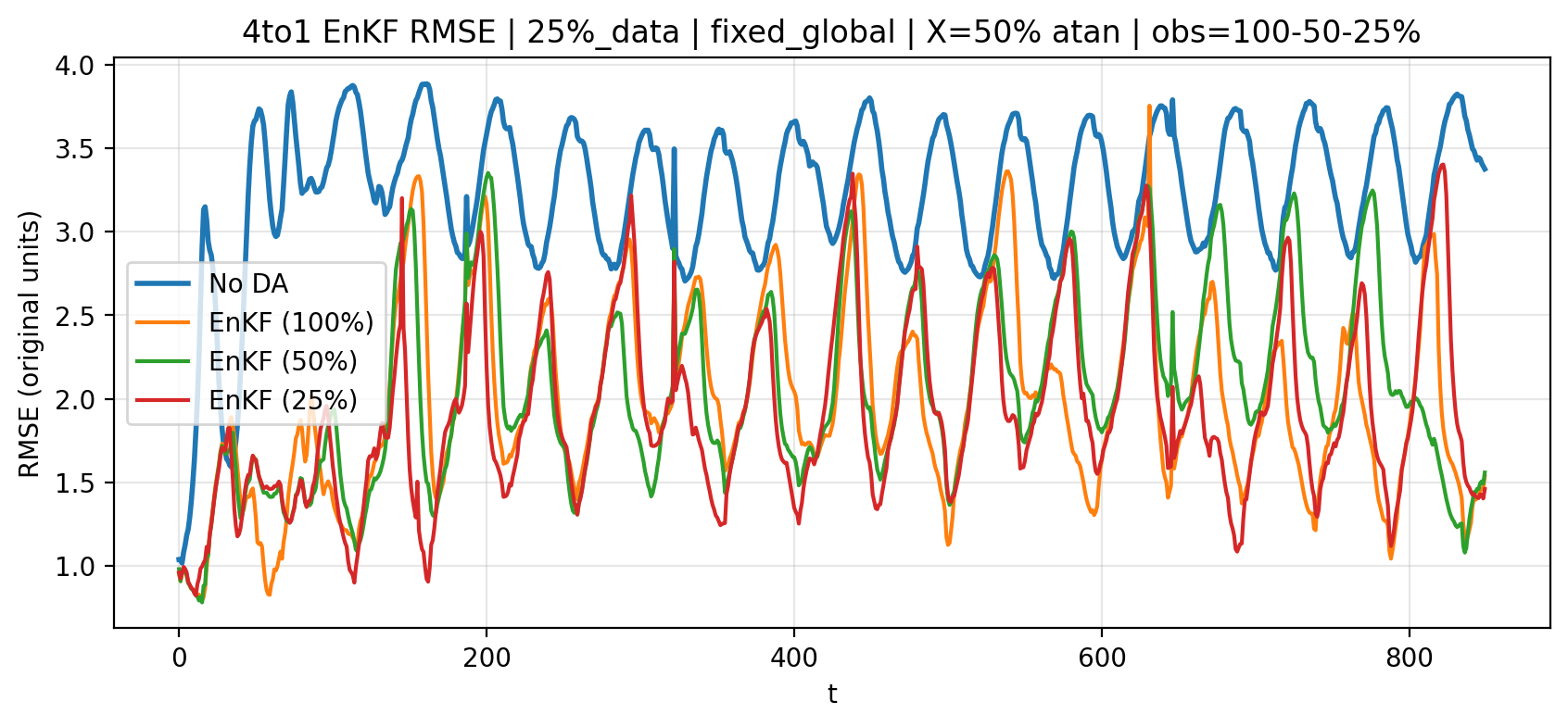}
        \caption{EnKF correction.}
        \label{fig:example3_rmse_enkf}
    \end{subfigure}
    \caption{RMSE comparison between EnSF and EnKF for the $4$-to-$1$ model trained with the reduced training split under mixed direct/arctangent observations. The no-DA curve corresponds to the open-loop forecast, while the data-assimilation curves correspond to correction using $25\%$, $50\%$, and $100\%$ observation levels. The RMSE is computed over all $5000$ state components.}
    \label{fig:example3_rmse}
\end{figure}

Fig.~\ref{fig:example3_rmse} compares the RMSE curves obtained by EnSF and EnKF. The RMSE values are computed over all $5000$ state components after inverse normalization. Both methods reduce the error relative to the no-DA open-loop forecast, confirming again that sequential observational correction is essential in this setting. However, the EnSF-corrected curves remain substantially below the no-DA curve and exhibit relatively low RMSE throughout the time interval. In contrast, although EnKF also improves upon the open-loop forecast, its RMSE remains much larger than that of EnSF for most of the experiment. This indicates that the EnSF update provides a more effective correction mechanism for this nonlinear high-dimensional filtering problem.

\begin{figure}[p]
    \centering
    \begin{subfigure}[t]{0.49\textwidth}
        \centering
        \includegraphics[width=\textwidth]{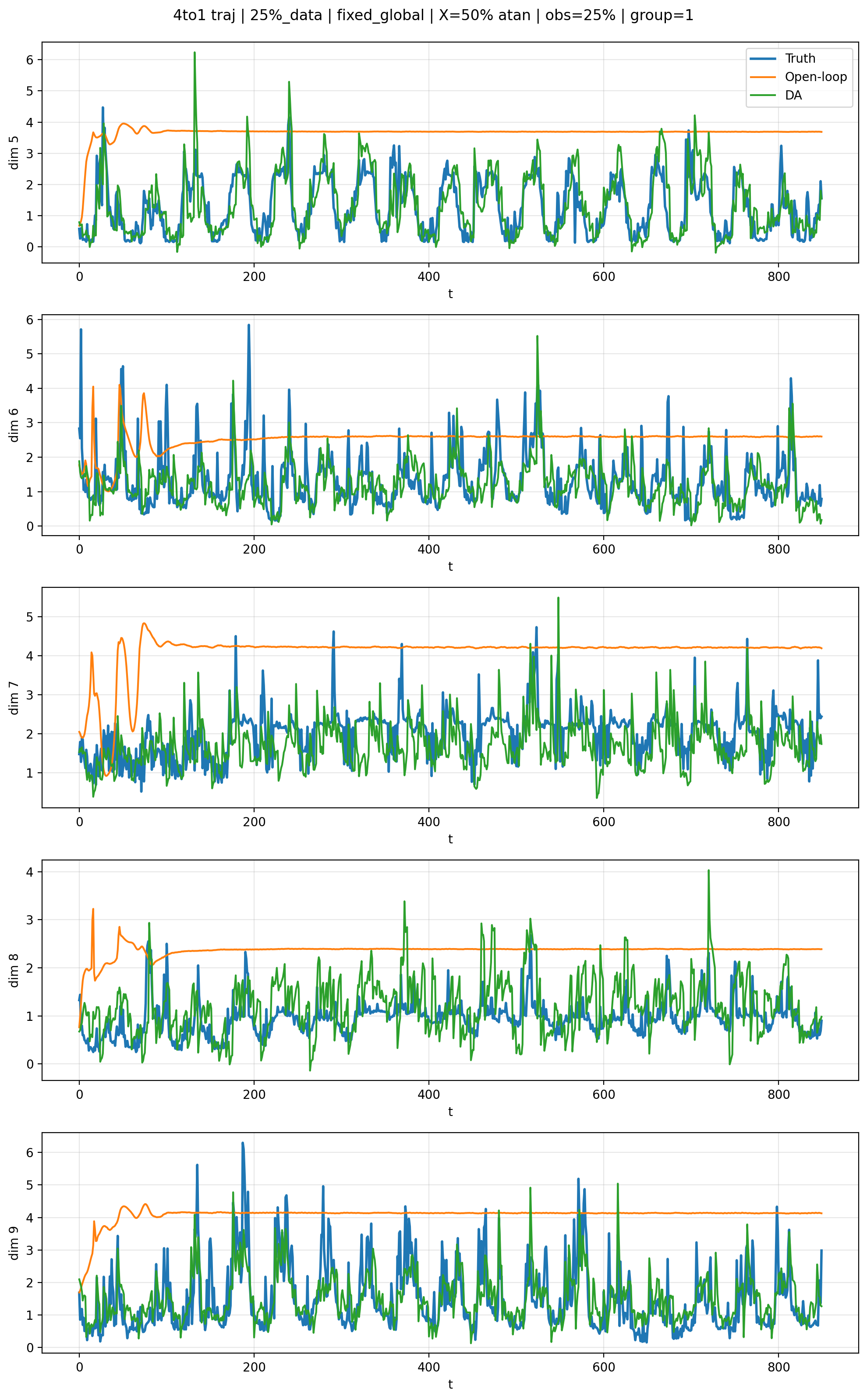}
        \caption{EnSF correction with $25\%$ observations.}
        \label{fig:example3_traj_ensf}
    \end{subfigure}
    \hfill
    \begin{subfigure}[t]{0.49\textwidth}
        \centering
        \includegraphics[width=\textwidth]{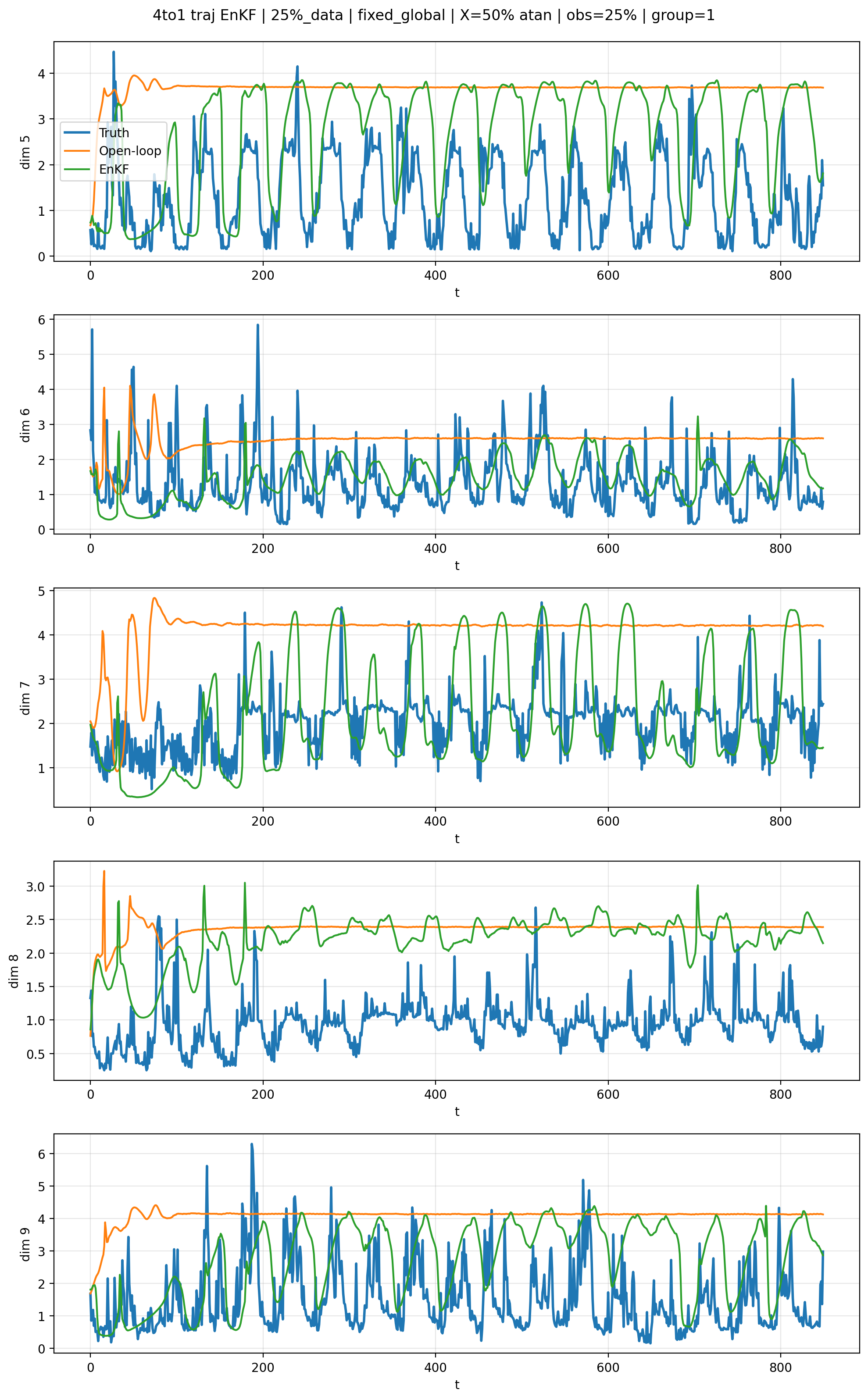}
        \caption{EnKF correction with $25\%$ observations.}
        \label{fig:example3_traj_enkf}
    \end{subfigure}
    \caption{Representative trajectory comparison between EnSF and EnKF for the $4$-to-$1$ model trained with the reduced training split under mixed direct/arctangent observations. The EnSF-corrected trajectory more closely follows the truth, while the EnKF result shows weaker correction in the displayed dimensions.}
    \label{fig:example3_traj}
\end{figure}

To further compare the two methods at the trajectory level, Fig.~\ref{fig:example3_traj} shows representative trajectories using $25\%$ observations. The displayed trajectories correspond to selected representative state dimensions. The blue curves denote the truth, the orange curves denote the open-loop forecast, and the green curves denote the data-assimilation result. In the EnSF case, the corrected trajectory follows the main temporal structures of the truth across the displayed dimensions. The open-loop forecast, by contrast, remains far from the truth after the initial transient. In the EnKF case, the corrected trajectory improves upon the open-loop forecast but does not track the truth as consistently as EnSF. In several displayed dimensions, EnKF produces overly smooth or displaced trajectories, indicating that the Kalman-type correction is less effective for this mixed nonlinear observation setting.

Overall, this example demonstrates the advantage of EnSF over EnKF for the forecast-correction task considered here. While both methods use the same learned forward model and the same observational information, EnSF achieves a stronger reduction in RMSE and better trajectory recovery. This result is consistent with the motivation for using a score-based ensemble filter: the filtering distribution in this problem can be nonlinear and non-Gaussian, and the EnSF update provides a more flexible correction than the Gaussian update structure used by EnKF.

\FloatBarrier
\section{Conclusion}

In this work, we used the Ensemble Score Filter to improve state estimation for real-data energy-consumption forecasting. The forward dynamics were represented by a pretrained spatio-temporal forecasting model, which was treated as a black-box state propagator. Through numerical experiments, we showed that the pretrained model can provide reasonable short-horizon predictions when correct input information is available, but its open-loop forecasts may become unreliable over long horizons. By assimilating partial and noisy observations, EnSF substantially reduced the forecast error and recovered the main temporal features of the true consumption trajectories. The improvement was observed for both insufficiently trained and fully trained $4$-to-$1$ forecasting models.

We also compared EnSF with EnKF under a mixed direct/nonlinear observation setting. The results show that EnSF achieves lower RMSE and better trajectory recovery than EnKF in the considered high-dimensional filtering problem. Overall, these results demonstrate the potential of EnSF as an effective data assimilation tool for correcting learned energy-consumption forecasts. Future work may extend the current framework by incorporating additional exogenous variables, especially temperature, into the forecasting model so that power consumption can be predicted and corrected under weather-dependent demand patterns. It would also be useful to validate the framework on additional real-world energy datasets and to study more complex observation patterns and missing-data mechanisms.

\bibliographystyle{elsarticle-num}
\bibliography{reference}

\end{document}